\documentclass{article}

\usepackage{amsmath,amsfonts}
\usepackage{algorithmic}
\usepackage{algorithm}
\usepackage{array}
\usepackage[caption=false,font=normalsize,labelfont=sf,textfont=sf]{subfig}
\usepackage{textcomp}
\usepackage{stfloats}
\usepackage{url}
\usepackage{verbatim}
\usepackage{graphicx}
\usepackage{cite}
\usepackage{hyperref}
\usepackage{booktabs}
\usepackage{orcidlink}
\usepackage{microtype}
\usepackage{float} 
\usepackage{adjustbox} 
\usepackage{xcolor} 
\usepackage[margin=1in]{geometry}

\title{Online Continual Learning: A Systematic Literature Review of Approaches, Challenges, and Benchmarks}

\author{
    Seyed Amir Bidaki\,$^{\dagger}$\ \orcidlink{0009-0003-4537-0910}, 
    Amir Mohammadkhah\,$^{\dagger}$\ \orcidlink{0009-0001-3697-8057}, 
    Kiyan Rezaee\,$^{\dagger}$\ \orcidlink{0009-0006-9922-8188},\\
    Faeze Hassani\,\orcidlink{0009-0002-9212-7676}, 
    Sadegh Eskandari\,\orcidlink{0000-0002-3007-2369},
    Maziar Salahi\,\orcidlink{0000-0002-0430-3063}, \\
    and Mohammad M. Ghassemi\,\orcidlink{0000-0001-5135-8588}
}

\date{} 

\begin{document}

\maketitle

\begin{abstract}
Online Continual Learning (OCL) is a critical area in machine learning, focusing on enabling models to adapt to evolving data streams in real-time while addressing challenges such as catastrophic forgetting and the stability-plasticity trade-off. This study conducts the first comprehensive Systematic Literature Review (SLR) on OCL, analyzing 81 approaches, extracting over 1,000 features (specific tasks addressed by these approaches), and identifying more than 500 components (sub-models within approaches, including algorithms and tools). We also review 83 datasets spanning applications like image classification, object detection, and multimodal vision-language tasks. Our findings highlight key challenges, including reducing computational overhead, developing domain-agnostic solutions, and improving scalability in resource-constrained environments. Furthermore, we identify promising directions for future research, such as leveraging self-supervised learning for multimodal and sequential data, designing adaptive memory mechanisms that integrate sparse retrieval and generative replay, and creating efficient frameworks for real-world applications with noisy or evolving task boundaries. By providing a rigorous and structured synthesis of the current state of OCL, this review offers a valuable resource for advancing this field and addressing its critical challenges and opportunities. The complete SLR methodology steps and extracted data are publicly available through the provided link: \url{https://github.com/kiyan-rezaee/Systematic-Literature-Review-on-Online-Continual-Learning}
\end{abstract}


\section{Introduction}

The growing integration of Artificial Intelligence (AI) across industries underscores the need for models capable of adapting to dynamic and evolving environments. Continual learning (CL) is a sub-field of machine learning designed to enable models to learn and update their knowledge incrementally from a continuous stream of data~\cite{lesort2020continual, vision, 10341211, language, Shaheen2022-fj}. The ability of CL  models is essential for developing models that are adaptable and resilient, capable of managing changing environments while avoiding the loss of previously learned knowledge—a challenge commonly known as catastrophic forgetting~\cite{10444954, COSSU2021607, hadsell2020embracing}.

Online Continual Learning (OCL) builds on CL by emphasizing the ability to learn from data streams in \textbf{real-time}~\cite{bonicelli2023effectiveness, Vodisch_2023_CVPR}. This approach closely resembles human cognition, as it gradually acquires and refines knowledge over time. OCL is effective for addressing real-world challenges in dynamic and unpredictable environments, such as autonomous driving, end-to-end speech processing, and robotics~\cite{verwimp2023continual, yang2022online}. Similar to CL, a key challenge in OCL models is addressing catastrophic forgetting.  To tackle this, researchers have proposed various methods. Experience replay methods~\cite{lin1992self} save and revisit past data to reinforce previously learned knowledge. Knowledge distillation~\cite{hinton2015distillingknowledgeneuralnetwork} also helps by transferring knowledge from the older model to the updated one, which supports maintaining performance on earlier tasks.

Interest in OCL has grown considerably in recent years, as evidenced by the increasing number of publications on the topic. However, despite this increased attention, a comprehensive study providing a clear and holistic overview of all proposed approaches in this field is still lacking. While some review studies on OCL have been conducted, their focus has remained relatively narrow.

Parisi et al.\cite{parisi2020online} provide a comprehensive overview of foundational concepts in OCL, such as catastrophic forgetting and the plasticity-stability trade-off. They also discuss biologically inspired strategies, which draw insights from how the brain learns and adapts over time. To address the challenges posed by sequential, non-i.i.d. data, they classify existing approaches into three main categories: synaptic-based methods, which focus on protecting important parameters through techniques like synaptic regularization; structural adaptation strategies, which involve dynamically modifying network architecture, such as adding or pruning neurons and connections; and self-organizing neural network models, which are inspired by biological networks and focus on organizing and adapting their structure in response to new data. Similarly, Hayes et al.\cite{hayes2022online} investigate OCL in the context of constrained devices, focusing on memory and computational limitations while assessing sample efficiency and inference methods across various CNN architectures. Mai et al.~\cite{mai2022online} offer an empirical analysis of OCL techniques, emphasizing the management of forgetting in class and domain incremental settings\footnote{Class-Incremental Learning entails learning new classes while retaining knowledge of previous ones, whereas Domain-Incremental Learning emphasizes generalizing across sequentially introduced domains for the same task without domain or task identity information during inference. Additional details can be found in Section~\ref{definition}.} and discussing key metrics such as accuracy and knowledge transfer.

While these review studies offer valuable insights, they are limited in scope, focusing on specific methods or categories instead of providing a comprehensive overview of OCL as a whole. To address this gap, we conducted the \textbf{first SLR in OCL}, following Newman’s guidelines~\cite{Newman} to ensure a rigorous and structured synthesis of the current state of the field. Our methodology provides a comprehensive and unbiased synthesis of existing research, focusing on clearly defined inclusion criteria, data extraction protocols, and quality assessments~\cite{Lame_2019, guidetoSLR}.

In our work, we \textbf{analyze 81 unique OCL approaches}, identifying more than 500 components, including modular sub-models, algorithms, and tools, and extracting over 1,000 features, which includes task settings and strategies for overcoming OCL-specific challenges. Additionally, we \textbf{compile a comprehensive list of 83 datasets used in OCL literature}, highlighting the diverse applicability of OCL methods across multiple domains. This detailed review lays a strong foundation for future research and offers valuable insights into advancing the understanding of OCL.

Our contributions are summarized as follows:

\begin{enumerate}

\item We conducted the first SLR in OCL, adhering to Newman’s guidelines. Our structure SLR methodology ensures an unbiased and comprehensive analysis, enabling us to identify unresolved challenges, propose future research directions, and highlight opportunities to advance the field.

\item We performed an in-depth analysis of 81 OCL approaches, identifying over 500 components. We also extract more than 1,000 features, covering task settings and strategies for addressing OCL-specific challenges.

\item We compiled a detailed list of 83 datasets, offering a valuable resource for researchers working on OCL applications across various domains, from image classification to specialized areas like autonomous systems and robotics.

\end{enumerate}

The remainder of this paper is organized as follows. Section~\ref{methodology} outlines the systematic methodology, including the formulation of research questions, selection criteria, and data extraction procedures. Section~\ref{analysis} presents a comprehensive analysis of OCL approaches, datasets, and evaluation metrics, with a focus on categorizing strategies of approaches and identifying key components and challenges. In Section~\ref{discussion}, we address the research questions, discuss unresolved challenges, propose future research directions, and provide valuable resources to advance OCL research.

\section{Methodology}\label{methodology}

In this section, we outline our methodology, which is adapted from Newman’s framework and visually summarized in Figure~\ref{fig:Methodology}. A brief description of Newman’s methodology is included in Appendix Section~\ref{g}, while comprehensive details can be found in the original work~\cite{Newman}.

\begin{figure*}[htbp]
    \centering
    \includegraphics[width=0.90\textwidth]{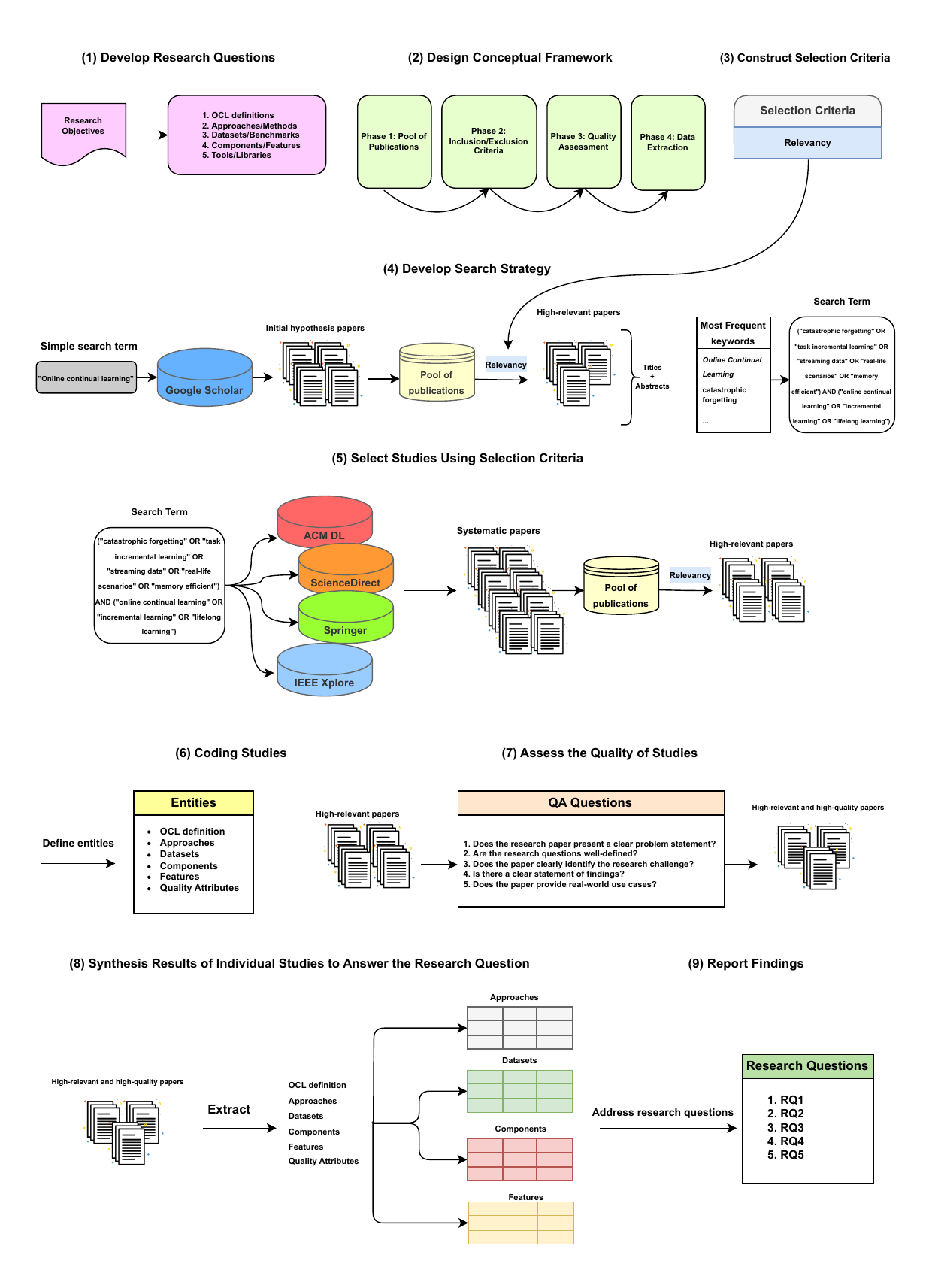}
    \caption{\footnotesize
        Our systematic methodology in this work. Our methodology starts with (1) defining research objectives, followed by developing research questions. (2) A conceptual framework is designed, and (3) selection criteria such as relevancy are constructed. (4) A search strategy is implemented using initial keywords and refined to extract more publications. (5) Studies are selected by applying the criteria to identify highly relevant papers. (6) The selected studies are coded to classify entities such as approaches, features, and quality attributes. (7) Quality assessment is performed to ensure high relevance and high-quality papers are included using defined questions. Finally, (8) extracted data are synthesized, and (9) findings are reported to address the research questions.}
    \label{fig:Methodology}
\end{figure*}

\subsection{Develop Research Questions}\label{develop} To comprehensively examine the OCL, we propose the following research questions. These questions are designed to explore the practical application of methods, the types of input data and benchmarks used, the common components and features of proposed approaches, as well as the tools and libraries employed the most in this field.

\begin{itemize}
\item \textbf{RQ1}: How is the problem setup defined and structured in OCL?
\item \textbf{RQ2}: How are various approaches and methods applied in practical situations?
\item \textbf{RQ3}: What datasets and benchmarks are typically employed to assess performance?
\item \textbf{RQ4}: What are the common components and features of the proposed approaches?
\item \textbf{RQ5}: What are the most prominent tools and libraries in this field? \end{itemize}

\subsection{Design Conceptual Framework} In this step, we designed our conceptual framework, which outlines the process from constructing selection studies (Subsection~\ref{construct}) to synthesizing results (Subsection~\ref{synthesis}). The purpose of designing the conceptual framework is to facilitate the subsequent stages of the methodology. Our conceptual framework consists of four phases, with the number of research papers included in each phase illustrated in Table~\ref{N_P}.

\begin{figure}     \centering     \includegraphics[width=0.8\linewidth]{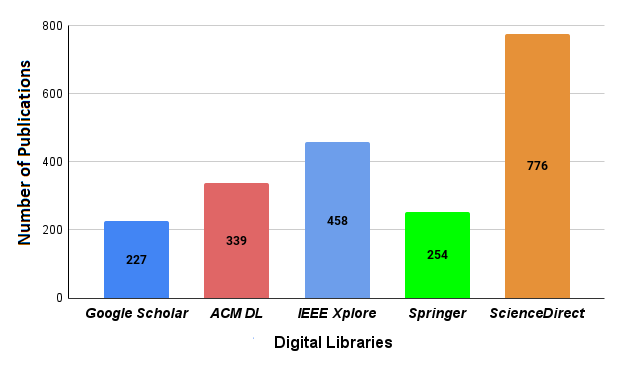}     \caption{Distribution of articles extracted from multiple digital libraries. This chart depicts the number of OCL-related studies retrieved from various sources, highlighting ScienceDirect as the largest contributor, with 776 articles from a total of 2061.}     \label{NUMpapers_phase1} \end{figure}

\begin{itemize}

\item \textbf{Phase 1: Pool of Publications} - This phase focused on gathering a diverse set of relevant studies from multiple sources, including (1) Google Scholar, (2) ACM Digital Library, (3) ScienceDirect, (4) Springer, and (5) IEEE Xplore. Figure~\ref{NUMpapers_phase1} presents the number of research papers collected from each source. The process is further explained in detail in subsections~\ref{construct} and~\ref{search}.

\item \textbf{Phase 2: Inclusion/Exclusion Criteria} - In this phase, we implemented specific criteria to refine the collected studies, selecting those most relevant to our research questions and excluding those that did not meet the established standards. This process is further detailed in subsection~\ref{select}.

\item \textbf{Phase 3: Quality Assessment} - This phase involved evaluating the quality of the selected studies, with a focus on their methodologies, to ensure our analysis is grounded in reliable and robust research. Details are provided in subsection~\ref{assess}.

\item \textbf{Phase 4: Data Extraction} - This phase focused on extracting and synthesizing data from highly relevant and high-quality studies, providing the foundation for a comprehensive analysis aligned with our research objectives. For details, refer to subsections~\ref{coding} and~\ref{synthesis}.

\end{itemize}

\subsection{Construct Selection Criteria}\label{construct}

To identify research papers highly relevant to OCL, we adopted a systematic approach that combines keyword extraction with semantic similarity analysis. At this stage, suppose a pool of research papers ($P$) were collected from multiple digital libraries (further details are provided in subsection~\ref{search}). The goal was to develop criteria to filter this pool and include only high-relevant studies. The process was carried out as follows: 

\begin{enumerate}  
    \item \textbf{Data Collection}:  
    We retrieved research papers from the \textbf{OpenAlex}~\cite{priem2022openalex} and \textbf{arXiv} databases using two targeted search terms:  
    \begin{itemize}  
        \item \textit{``Online Continual Learning"}: This search captured studies specifically focused on OCL, which we refer to as (\textbf{$A$}).
        \item \textit{``Continual Learning"}: This search gathered a broader range of CL and its variants, which we refer to as (\textbf{$B$}).
    \end{itemize}  

    \item \textbf{Keyword Extraction}:  
    From the retrieved papers ($A$) and ($B$), we extracted key terms that encapsulated the core concepts of each study. These keywords were categorized as follows:  
    \begin{itemize}  
        \item \textbf{$K_1$}: A set of keywords representing the research papers in \textbf{$A$}.  
        \item \textbf{$K_2$}: A set of keywords representing the research papers in \textbf{$B$}.
    \end{itemize}  

    \item \textbf{Categorization of Publications}:  
    Using the extracted keyword sets, we assessed the relevance of each paper in our pool ($P$) by calculating its semantic similarity to both \textbf{$K_1$} and \textbf{$K_2$}. Based on these similarity scores, each paper was categorized into one of the following three groups:  
    \begin{itemize}  
        \item \textbf{Low Relevance}: Papers with a similarity score lower than \textbf{0.5} for both \textbf{$K_1$} and \textbf{$K_2$}, indicating limited relevance to the CL and its variants, including OCL.
        
        \item \textbf{Medium Relevance}: Papers with a similarity score higher than \textbf{0.5} to \textbf{$K_2$}, but lower than \textbf{0.5} to \textbf{$K_1$}, indicating a focus on CL or its variants, but with less specific alignment to OCL.

        \item \textbf{High Relevance}: Papers with a similarity score higher than \textbf{0.5} to \textbf{$K_1$}, indicating a strong focus on OCL-specific methodologies and concepts.
        
    \end{itemize}  
\end{enumerate}  

We used \textbf{Sentence-BERT (SBERT)}~\cite{reimers2019sentence} as the embedding model to calculate the semantic similarity between the papers and the defined keyword sets. Research papers categorized as ``High Relevance" were selected for the subsequent steps.

\subsection{Develop Search Strategy}\label{search}

We initiated the search process with a simple search term, ``Online Continual Learning", extracting 227 of the most relevant research papers from Google Scholar search results. This initial set was labeled as the ``Initial Hypothesis". Using this set, we refined our search terms through the following steps:

\begin{enumerate}
\item Identify high-relevance research papers from the initial set based on the selection criteria outlined in Section~\ref{construct}.
\item Extract the titles and abstracts of the selected papers and compile all words into a collective pool.
\item Determine the most frequent words in this pool, excluding common stopwords.
\item Formulate a refined search term by connecting the most frequent words using the ``OR" and ``AND"  boolean operators.
\end{enumerate}

We used the ``AND" operator at the end to ensure that at least one of the terms ``online continual learning", ``incremental learning", or ``lifelong learning" was included in the extracted papers. The final search terms were: \\
\textbf{(``catastrophic forgetting" OR ``task incremental learning" OR ``streaming data" OR ``real-life scenarios" OR ``memory efficient") AND (``online continual learning" OR ``incremental learning" OR ``lifelong learning")}

Using the final refined search term, we retrieved 1,834 research papers from digital databases, including ACM DL, IEEE Xplore, ScienceDirect, and Springer, to ensure comprehensive coverage. This collection of articles was labeled as the ``Systematic" set.

\subsection{Select Studies Using Selection Criteria}\label{select}
Two groups of papers were extracted: (1) ``Initial hypothesis", which includes papers manually extracted first from Google Scholar, and (2) ``Systematic", which includes papers systematically extracted from digital databases like ACM DL, IEEE Xplore, ScienceDirect, and Springer using the constructed search term. Using the selection criteria mentioned in Section~\ref{construct}, we include 170 highly relevant studies for the next step.

\begin{table}[t!]
    \centering
    \caption{Publication counts at each phase of our conceptual framework.}
    \begin{tabular}{lc}
        
        \toprule
        \textbf{Phase} & \textbf{Publication Count} \\
        \midrule
        \textbf{Phase 1: Pool of Publications} & 2061 \\
        \textbf{Phase 2: Inclusion/Exclusion Criteria} & 170 \\
        \textbf{Phase 3: Quality Assessment} & 170 \\
        \textbf{Phase 4: Data Extraction} & 81 \\
        \bottomrule
    \end{tabular}
    
    \label{N_P}
\end{table}

\subsection{Coding Studies}\label{coding} In this step, we defined the entities we aim to extract from each research paper to answer our research questions: (1) the definition of OCL (if provided); (2) approaches and methods; (3) datasets; (4) components and features; and (5) tools and libraries used in the studies. To ensure consistency, all entities except the first—which is self-explanatory—were clearly defined before extraction began.

\begin{itemize}     \item \textbf{Approaches/Methods} are the techniques, algorithms, or procedures used in the research to solve the problem or address the research question. It includes the overall strategy, such as machine learning algorithms, statistical analysis, or experimental setups, which are applied to the data or problem domain.      \item \textbf{Datasets} are collections of the data used in the study for analysis, training, and validation. Datasets may include raw data, processed data, or any other form of data utilized by researchers.      \item \textbf{Components} are the specific elements used within the approaches/methods to achieve the research objectives. Components may include algorithms, frameworks, or systems integrated into the overall methodology.  

\item \textbf{Features} are the unique characteristics or attributes of a component that distinguish it from other components. These could include aspects such as the model architecture, parameters, learning rate, or any other defining traits that set it apart from other models within the same or different approaches/methods.     \item \textbf{Quality attributes} are the criteria, features, or metrics that qualitatively describe an approach or its components. \end{itemize}   

\subsection{Assess the Quality of Studies}\label{assess} After applying the selection studies criteria, we identified 170 papers relevant to our study. However, not all of these studies met the quality standards required for a thorough review and data extraction. To address this, we excluded low-quality research papers before proceeding with data extraction. We evaluated each paper using five yes/no questions to assign a value and determine its suitability for review:

\begin{enumerate} \item Does the research paper present a clear problem statement? \item Are the research questions well-defined? \item Does the paper clearly identify the research challenge? \item Is there a clear statement of findings? \item Does the paper provide real-world use cases? \end{enumerate}

We considered research papers with at least three ``yes" answers as high-quality articles, making them suitable for data extraction.   

\subsection{Synthesis Results of Individual Studies to Answer the Research Question}~\label{synthesis} We synthesized and transformed the findings from individual studies to provide a comprehensive answer to the research questions. We quantified the occurrence of all defined entities across the included studies to identify areas of consensus and frequent focus. This approach offered a quantitative perspective, highlighting the significance of each entity based on its prevalence in the literature. For example, the analysis identified common strategies for mitigating catastrophic forgetting and emphasized the tasks and settings most frequently addressed by OCL approaches. This framing of dominant themes in current research helps to illustrate which entities have gained widespread acceptance. Additionally, our method uncovered under-researched entities, revealing gaps or emerging opportunities in OCL studies that merit closer attention.  

\subsection{Report Findings}~\label{report}

Finally, we present our findings by synthesizing the collected data and addressing each research question in detail. To address RQ1, we offer a unified definition of OCL, providing clarity and consistency for researchers in this field, as discussed in Section~\ref{definition}.

For RQ2, we identified and categorized 81 approaches proposed within the OCL literature into three main strategy types: Replay-based, Architecture-based, and Regularization-based. Each category provides unique mechanisms for addressing the challenges of OCL, offering diverse perspectives on solving these issues, which are further elaborated in Section~\ref{Approaches}.

In response to RQ3, we analyzed 83 datasets and benchmarks employed across OCL studies. This analysis highlights the extensive range of real-world applications supported by OCL approaches, reflecting the practical relevance of this research area, as detailed in Section~\ref{Datasets}.

For RQ4, we investigated the components and features frequently incorporated within OCL approaches. Our analysis indicates that these components often include modular architectures, memory buffers, and feature extraction techniques, all designed to enhance learning efficiency and adaptability, as explained in Sections~\ref{Components} and~\ref{Features}.

Lastly, in addressing RQ5, we explored the tools and libraries commonly used in OCL research. Our findings reveal that general-purpose libraries such as PyTorch and TensorFlow, along with specialized frameworks like Avalanche, are instrumental in supporting the implementation and evaluation of OCL approaches, as discussed in Section~\ref{Components}.

\section{Results and Analysis}\label{analysis}

In this section, we provide an analysis of the data extracted from the 81 selected research papers. We start with a unified definition of OCL in Section~\ref{definition}. Then, in Section~\ref{Approaches}, we categorize the extracted approaches according to their underlying strategies. In Section~\ref{Components}, we examine the key components of each approach, focusing on model architectures, optimization techniques, and loss functions. Section~\ref{Features} follows, where we analyze the unique features of all approaches, organizing them by type and purpose. After that, in Section~\ref{QA}, we examine the quality attributes that describe these approaches and discuss how changes in these attributes affect their objectives. Section~\ref{Datasets} reviews the datasets used in the selected studies. Finally, in Section~\ref{metrics}, we list the key metrics in this field and present their corresponding mathematical formulas.

\subsection{OCL Definition}\label{definition}

One of the gaps in the OCL field is the lack of a unified definition for the OCL setup. To address this, we reviewed the selected research papers and collected the definitions of OCL provided by the authors, where available. In this section, we summarize the extracted information, including the formal formulation of CL, its variants (see Table~\ref{tab:continual_learning_scenarios}), and the specific details of OCL.

\subsubsection{CL Formulation}

CL is characterized as a process where a model, denoted as $\theta$, learns from a sequential stream of tasks or data distributions. A fundamental aspect of CL is the model's ability to adapt to new tasks while potentially having limited or no access to past training data. Formally, when a new batch of training data corresponding to a task $t$ is introduced, it can be represented as:

$ D_{t,b} = \{X_{t,b}, Y_{t,b}\} $
where:
\begin{itemize}
    \item $X_{t,b}$ represents the input data,
    \item $Y_{t,b}$ denotes the corresponding labels,
    \item $t \in T = \{1, 2, \ldots, k\}$ indicates the task identity,
    \item $b \in B_t$ signifies the index of the batch.
\end{itemize}

The CL model must effectively learn from these dynamic data distributions while maintaining performance across all tasks~\cite{10444954}. Building on this formulation, we explore the diverse scenarios within CL that contextualize its application and challenges.

\subsubsection{CL Variants}

CL contains diverse scenarios, each characterized by the nature of task identities and the distribution of training samples. One such scenario is \textbf{Instance-Incremental Learning (IIL)}, where all training samples belong to the same task~\cite{lomonaco2017core50}. For example, a model initially trained on a dataset of cat images might later receive additional images of cats in different poses or lighting conditions, incrementally expanding its knowledge.

In contrast, \textbf{Domain-Incremental Learning (DIL)} deals with tasks that share the same label space but differ in input distributions~\cite{hsu2018re, van2019three}. For instance, a model trained on handwritten digit images might subsequently encounter real-world images of digits, requiring adaptation to the new input distribution.

\textbf{Task-Incremental Learning (TIL)} differs further by involving distinct data label spaces, where task identities are explicitly provided during both training and testing. For example, a model trained to classify animals might later learn to classify vehicles and plants, with task identities guiding the learning process. Similarly, \textbf{Class-Incremental Learning (CIL)} involves multiple tasks, but here, task identities are only available during training. The model learns from ($ {D_{t,b}, t}) $ while performing classification across all classes simultaneously~\cite{hsu2018re, van2019three}. For example, a model trained to classify dogs and cats might later learn to classify birds, ensuring it retains knowledge of the previous classes.

Moving beyond task-defined settings, \textbf{Task-Free Continual Learning 
  (TFCL)} eliminates the provision of task identities, focusing instead on disjoint label spaces~\cite{aljundi2019task}. For example, a model might first classify animal images and later classify object images without any explicit task boundaries, adapting dynamically as data arrives. On the other hand, \textbf{Online Continual Learning (OCL)} emphasizes real-time learning by processing tasks with disjoint label spaces through a one-pass data stream~\cite{aljundi2019gradient}. 

Additional challenges arise in \textbf{Blurred Boundary Continual Learning (BBCL)}, where data label spaces overlap, making it harder to differentiate tasks~\cite{bang2021rainbow, buzzega2020dark}. For example, a model trained to classify different bird species might encounter classes with highly similar features, such as sparrows and small birds, complicating task distinctions. Lastly, \textbf{Continual Pre-training (CPT)} involves sequentially arriving pre-training data, enabling gradual knowledge transfer without explicit task boundaries~\cite{sun2020ernie}.

The diversity of CL scenarios highlights the varying challenges models face, ranging from managing task identities to handling overlapping or disjoint label spaces. Among these, \textbf{OCL} stands out due to its unique emphasis on real-time adaptation and stringent constraints on data usage. In OCL, the focus shifts from accommodating explicit task structures to navigating dynamic, single-pass learning environments, as detailed below.

\subsubsection{OCL}

Working in an OCL setup typically requires meeting three key conditions: (1) data must be fed to the model in real-time and in small batches, meaning the model processes data as it becomes available and updates incrementally rather than accessing the entire dataset at once; (2) tasks at different time steps must have disjoint label spaces, indicating that the labels or categories of data from one task do not overlap with those from another task encountered later; and (3) data for each task is trained in a single epoch, with no opportunity to revisit previous data. However, some studies have not considered the second condition, choosing not to enforce disjoint label spaces~\cite{Vodisch_2023_CVPR, han2022selecting, kwon2022toward}. In OCL, task identity is optionally available. This means that there are no strict requirements for specifying task identity during either training or testing. Instead, its inclusion depends on the specific problem being addressed~\cite{10444954}. The defining characteristics of OCL include the following:

\begin{itemize} \item \textbf{Training Stream Representation}: The data is processed as a continuous stream, represented as $\vec{D} = ((x_t, y_t)){t=1}^T$, where each $(x_t, y_t)$ corresponds to the input-output pair at time step $t$. \item \textbf{Model Update Mechanism}: At each time step, the model undergoes an update using a functional transformation, $f(\theta_t, (x_t, y_t)) \to \theta_{t+1}$. Here, $\theta_t$ represents the model's state at time $t$, and $(x_t, y_t)$ is the current data instance being processed. \end{itemize}

Unlike \textbf{Offline Continual Learning}, which processes data in distinct sessions and allows models to revisit all samples within a task, OCL requires real-time adaptation with no opportunity to review previous data. This sequential nature of OCL introduces complexities, such as handling non-stationary data distributions and balancing new learning with the retention of past knowledge~\cite{son2024meta}.

The \textbf{task-free OCL} setup represents a more advanced type of OCL, adding complexity by removing clear task boundaries. In this scenario, models must adapt to a continuous stream of data without knowing when a new task begins, requiring flexibility in adjusting to changing data distributions while avoiding catastrophic forgetting. This is in contrast to the \textbf{class-incremental setting}, where new classes are learned one after another in isolation from the others~\cite{gu2024summarizing}.

Additionally, OCL methods generally do not store large amounts of data or exemplars for replay due to the memory and computational constraints inherent in online systems. Instead, exemplar-free methods, like those described in~\cite{he2022exemplar}, focus on learning directly from the data stream without relying on additional memory buffers. This aligns with the real-world requirement of limited storage and computational resources.

Another challenge in OCL is the potential for \textbf{non-IID (Independent and Identically Distributed)} data, where the distribution of data may change over time. OCL methods must, therefore, be robust to such distribution shifts and capable of learning effectively in dynamic environments, as demonstrated in~\cite{bonicelli2023effectiveness}.

A practical example of the OCL setup is seen in the OAK dataset, which mimics real-world scenarios through egocentric video streams collected over nine months. The dataset focuses on recognizing and learning new objects as they appear in daily routines, such as outdoor objects labeled across 105 categories.  OAK provides a situation for models to adapt incrementally by training and evaluating simultaneously on a data stream, addressing challenges like catastrophic forgetting. This setup mirrors dynamic environments, where models encounter new tasks, such as recognizing previously unseen object categories while retaining prior knowledge~\cite{wang2021wanderlust}.

\begin{table}[t!]
    \centering
    \caption{Overview of Continual Learning Scenarios}
    \begin{tabular}{@{}>{\raggedright}p{4.5cm} p{4.5cm} p{3.75cm}@{}}
        \toprule
        \textbf{Scenario} & \textbf{Task Identity} & \textbf{Label Space} \\ 
        \midrule
        Instance-Incremental Learning (IIL) & Not required & Single \\ 
        \midrule
        Domain-Incremental Learning (DIL) & Not required & Shared \\ 
        \midrule
        Task-Incremental Learning (TIL) & Provided during training and testing & Disjoint across tasks \\ 
        \midrule
        Class-Incremental Learning (CIL) & Provided only during training & Combined across tasks \\ 
        \midrule
        Task-Free Continual Learning (TFCL) & Not provided & Disjoint \\ 
        \midrule
        Blurred Boundary Continual Learning (BBCL) & Not provided & Overlapping \\ 
        \midrule
        Continual Pre-training (CPT) & Not applicable & Gradually expanding \\ 
        \midrule
        Online Continual Learning (OCL) & Optionally available & Disjoint \\ 
        \bottomrule
    \end{tabular}
    \label{tab:continual_learning_scenarios}
\end{table}

\subsection{Approaches}\label{Approaches}

\begin{figure*}[htbp]
    \centering
    \includegraphics[width=1\linewidth]{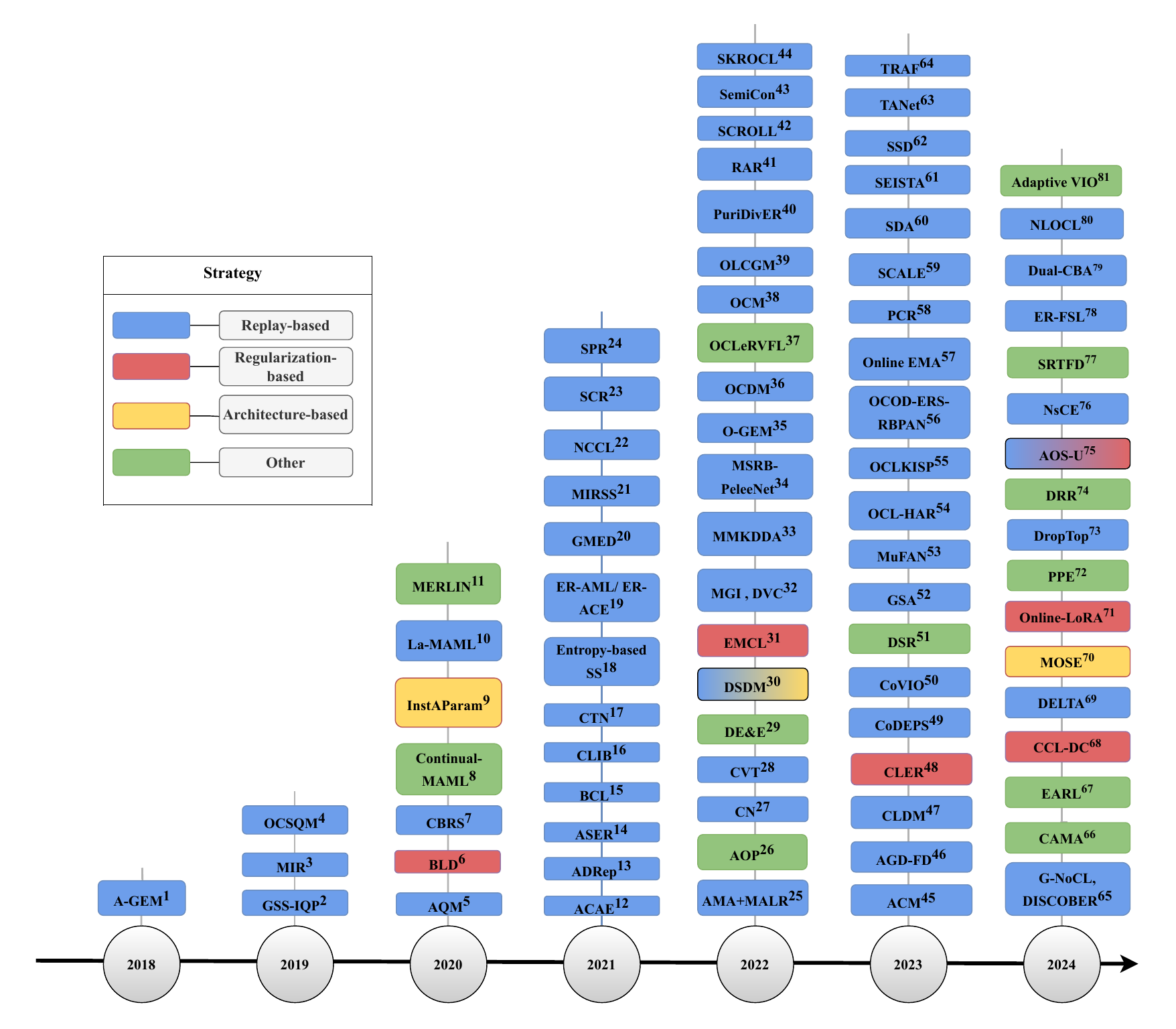}
    \caption{Classification of OCL approaches by strategy. The figure categorizes 81 identified approaches into replay-based, regularization-based, and architecture-based strategies, along with novel methods grouped as ``Other". Superscripts denote the reference numbers corresponding to each approach: 1.~\cite{chaudhry2018efficient}, 2.~\cite{aljundi2019gradient}, 3.~\cite{jiang2021multi}, 4.~\cite{caccia2019online}, 5.~\cite{caccia2020online}, 6.~\cite{fini2020online}, 7.~\cite{chrysakis2020online} 8.~\cite{caccia2020online}, 9.~\cite{NEURIPS2020_ca4b5656}, 10.~\cite{gupta2020look}, 11.~\cite{kj2020meta}, 12.~\cite{wang2021acae}, 13.~\cite{cai2021online}, 14.~\cite{shim2021online}, 15.~\cite{pham2020bilevel}, 16.~\cite{koh2021online}, 17.~\cite{pham2021contextual}, 18.~\cite{wiewel2021entropy}, 19.~\cite{caccia2021new}, 20.~\cite{jin2021gradient}, 21.~\cite{jiang2021multi}, 22.~\cite{yin2021mitigating}, 23.~\cite{mai2021supervised}, 24.~\cite{kim2021continual}, 25.~\cite{cai2022improving}, 26.~\cite{guo2022adaptive}, 27.~\cite{pham2022continualnormalizationrethinkingbatch}, 28.~\cite{10.1007/978-3-031-20044-1_36}, 29.~\cite{wojcik2022neural}, 30.~\cite{pourcel2022online}, 31.~\cite{zou2022efficient}, 32.~\cite{gu2022not}, 33.~\cite{HAN2022104966}, 34.~\cite{kwon2022toward}, 35.~\cite{yang2022online}, 36.~\cite{liang2022optimizing}, 37.~\cite{wong2022online}, 38.~\cite{pmlr-v162-guo22g}, 39.~\cite{sangermano2022sample}, 40.~\cite{bang2022online}, 41.~\cite{zhang2022simple}, 42.~\cite{wang2022schedule}, 43.~\cite{michel2022contrastivelearningonlinesemisupervised}, 44.~\cite{han2022selecting}, 45.~\cite{prabhu2023online}, 46.~\cite{michel2023learning}, 47.~\cite{xiao2023online}, 48.~\cite{bonicelli2023effectiveness}, 49.~\cite{vodisch2023codeps}, 50.~\cite{Vodisch_2023_CVPR}, 51.~\cite{huo2024non}, 52.~\cite{Guo_2023_CVPR}, 53.~\cite{jung2023new}, 54.~\cite{schiemer2023online}, 55.~\cite{Han_2023}, 56.~\cite{CHEN2023105549}, 57.~\cite{soutif2023improving}, 58.~\cite{lin2023pcr}, 59.~\cite{dam2022scalable}, 60.~\cite{yu2023mitigatingforgettingonlinecontinual}, 61.~\cite{harun2023siesta}, 62.~\cite{gu2023summarizing}, 63.~\cite{HONG2023126527}, 64.~\cite{kong2023trust}, 65.~\cite{seo2024just},
    66.~\cite{kim2024online}, 67.~\cite{seo2024learning}, 68.~\cite{wang2024improving}, 69.~\cite{raghavan2024delta}, 70.~\cite{yan2024orchestrate}, 71.~\cite{wei2024online}, 72.~\cite{li2024progressive}, 73.~\cite{kim2024adaptive}, 74.~\cite{lyu2024overcoming}, 75.~\cite{eeckt2024unsupervised}, 76.~\cite{wang2024forgetting}, 77.~\cite{zhao2024srtfd}, 78.~\cite{lin2024er}, 79.~\cite{wang2024dual}, 80.~\cite{cheng2024nlocl}, 81.~\cite{pan2024adaptive},
    respectively.}
    \label{fig:strategy}
\end{figure*}

We define an approach as a model, framework, or method designed to address challenges and achieve objectives in OCL. For instance, methods like MEFA~\cite{iscen2020memory}, which are part of the broader CL domain but not specifically designed for online settings, are outside the scope of this study. From the selected papers, we identified 81 approaches and their strategies (see Figure~\ref{fig:strategy}) that tackle key challenges in OCL, such as mitigating catastrophic forgetting~\cite{guo2022adaptive, parisi2020online}, while efficiently managing memory~\cite{michel2024learning, koh2021online, pham2020bilevel} and computational resources~\cite{harun2023siesta, dhar2021survey}.

\begin{figure*}[t!]
    \centering
    \includegraphics[width=1\textwidth]{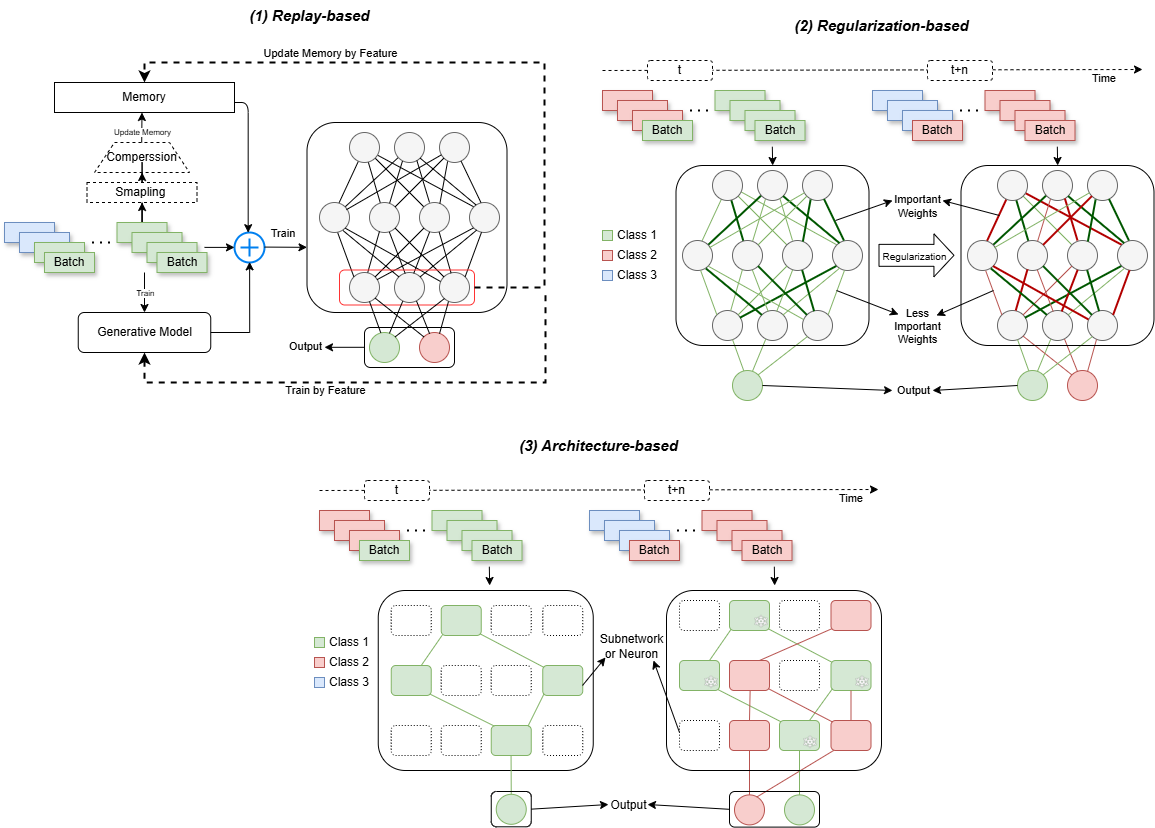}    \caption{Overview of key strategies in OCL. This figure illustrates three primary strategies in OCL: (1) Replay-based methods that store past data in memory or use generative models to sample and replay data during training, depicted by green and blue batches representing task data; (2) Regularization-based methods that preserve critical network weights, highlighted in green for important weights and red for less important weights, ensuring retention of prior knowledge; and (3) Architecture-based methods that dynamically allocate specific neurons or subnetworks to tasks, with color-coded neurons (green, red, and blue) representing task-specific knowledge while protecting previously learned information.}
    \label{fig:CoreS}
\end{figure*}

Most studies in this field categorize strategies into three main types: Replay-based, Architecture-based, and Regularization-based~\cite{parisi2020online, aljundi2019gradient, wong2022online, zou2022efficient, caccia2020online, kj2020meta}. Figure~\ref{fig:CoreS} provides a visual representation of the core ideas behind each of these strategies. One or more of these strategies can be combined to address challenges and achieve optimal performance across different settings. For instance, DSDM~\cite{pourcel2022online} used a combination of Replay-based and Architecture-based strategies.

\begin{enumerate}

    \item Replay-based methods:
    The central idea behind this strategy is to use a portion of previously encountered input data for replay, ensuring that the model retains what it has learned in later training phases~\cite{NEURIPS2020_ca4b5656}. A common approach is to use a limited buffer, called episodic memory, to store samples from past data streams and replay them later~\cite{zou2022efficient}, a technique known as Rehearsal. In this method, algorithms are used to select, compress, and merge samples from the input stream to optimize memory usage~\cite{sangermano2022sample}. A well-known example is Experience Replay (ER)~\cite{chaudhry2019tiny}, which trains the model on a combination of new input data and samples stored in episodic memory. However, storing raw data in memory can conflict with OCL settings, especially when data storage is restricted for privacy or security reasons. To address this, recent work~\cite{hayes2020remind} suggests storing learned representations instead of raw data, which greatly reduces memory costs.
    
    In addition to replay-based methods, generative models, known as pseudo-rehearsal, provide an alternative solution~\cite{wang2021acae, guo2022adaptive}. In this technique, a generative model is trained alongside the main model on incoming data. During later training phases, the generative model produces synthetic data that mimics previously seen data. The main model is then trained using a mix of new input data and generated data, helping it retain past knowledge. Pseudo-rehearsal reduces memory usage by avoiding direct storage of raw data but requires the added complexity of training a generative model. In contrast, while rehearsal methods rely on memory storage, they are simpler and often more effective in practice~\cite{zou2022efficient}.

    \item Regularization-based methods: These methods focus on minimizing changes to weights that are essential for retaining past knowledge while maintaining a fixed model capacity~\cite{kong2023trust, wojcik2022neural, zou2022efficient}. Although these strategies impose lower memory overhead, they often struggle to effectively mitigate catastrophic forgetting, especially in online settings and with long task sequences~\cite{wojcik2022neural, han2022selecting, aljundi2019online}.

    This approach can be divided into two categories: prior-focused and data-focused. The prior-focused approach aims to predict the importance of weights necessary for retaining previous knowledge, identifying those weights whose changes may lead to forgetting~\cite{wang2021acae, aljundi2019online}. To address this, a penalty is applied during optimization to limit changes to these critical weights~\cite{mai2022online}. In contrast, the data-focused approach uses knowledge distillation (KD)~\cite{wang2021acae, guo2022adaptive}, where a teacher-student mechanism transfers knowledge from the previous model to the new one~\cite{hinton2015distillingknowledgeneuralnetwork}. However, data-focused methods are vulnerable to domain shifts, and prior-focused methods may struggle to effectively constrain the optimization process, which can impact performance on earlier tasks~\cite{wang2021acae}.

    \item Architecture-based methods: These methods aim to address catastrophic forgetting by modifying the model’s structure and expanding its capacity~\cite{pourcel2022online, han2022selecting}. When new data with different distributions arrives, this strategy adapts by increasing the model’s capacity~\cite{han2022selecting}, adding task-specific parameters~\cite{zou2022efficient}, and assigning sub-networks for new tasks~\cite{kong2023trust}. This allows the model to retain previous knowledge while accommodating new learning~\cite{wang2021acae}. By isolating and freezing the parameters related to specific tasks, the remaining parameters can be trained for new knowledge, which is why this strategy is referred to as ``parameter isolation"~\cite{koh2021online, NEURIPS2020_ca4b5656}.

    Within this strategy, techniques such as pruning redundant weights~\cite{wong2022online}, using dynamically expandable networks (DEN)~\cite{yoon2017lifelong}, and employing Progressive Neural Networks (PNNs)~\cite{rusu2016progressive} are commonly used. Although these methods are effective in reducing catastrophic forgetting by increasing the model’s capacity, they face scalability challenges. As the number of tasks increases, the network must also expand to prevent forgetting, leading to significantly higher computational costs~\cite{han2022selecting}.

\end{enumerate}

As shown in Figure~\ref{fig:strategy}, replay-based methods have gained significant popularity, being used in 62 approaches. This popularity may be attributed to the scalability challenges and high computational costs associated with regularization-based and architecture-based strategies, which make them less practical for handling large and complex datasets~\cite{han2022selecting, wang2021acae, wojcik2022neural}.

\subsection{Components}\label{Components}

This section provides a comprehensive overview by categorizing key components, including structures, algorithms, and techniques across various approaches. We start with model architectures and optimization techniques, followed by essential elements like loss and activation functions. We then examine regularization methods, replay and memory buffer strategies, and data augmentation techniques. Lastly, we review classification and clustering methods alongside the libraries and frameworks commonly used in OCL implementations. A complete list of the extracted component entities and their corresponding approaches can be found in Appendix~\ref{componentsappendix}.

\begin{figure}
    \centering
    \includegraphics[width=0.9\textwidth]{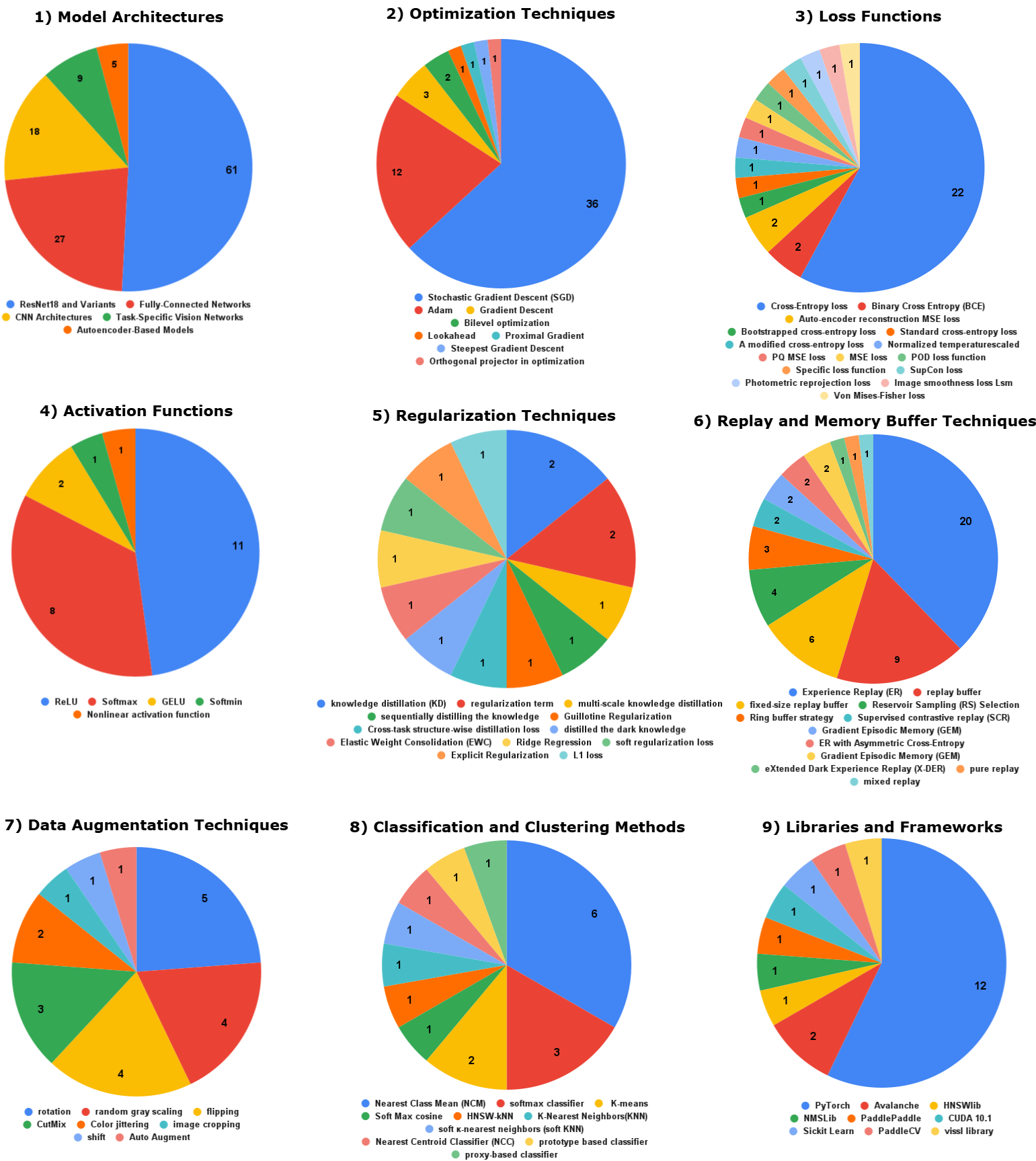}
    \caption{Prevalent components in OCL literature. The figure highlights frequently used components across 81 OCL studies, categorized by (1) Model architectures, (2) Optimization techniques, (3) Loss and activation functions, (4) Regularization strategies, (5) Replay and buffer techniques, (6) Data augmentation techniques, (7) Classification and clustering methods, and (8) Libraries and frameworks, with corresponding frequencies.}
    \label{fig:components}
\end{figure}

\subsubsection{Model Architectures} 

Among the extracted model architectures, \textbf{Convolutional Neural Networks (CNNs)} stand out as the most widely used. Among these, various CNN architectures have been employed in OCL approaches, with ResNet~\cite{he2016deep}—particularly \textbf{ResNet18}—emerging as the most prominent, appearing in 62 out of 81 approaches. The model’s type, depth, and size have a significant impact on learning performance. Therefore, selecting the right architecture is critical for tackling key challenges in OCL, such as catastrophic forgetting~\cite{lee2024impact}.

Researchers in this field used ResNet18 and its variants as common baselines due to several key advantages. One notable benefit is their shortcut connections, which help alleviate vanishing gradient issues and ensure stable learning in dynamic and evolving environments~\cite{he2016deep,lu2024revisitingneuralnetworkscontinual}. Compared to deeper models like ResNet-50, ResNet18 offers a practical balance between computational efficiency and accuracy, making it well-suited for tasks that require rapid adaptability without a substantial loss in performance~\cite{lee2024impact,chen2024redundancyfreesubnetworkscontinuallearning, Elharrouss_2024}.

Studies using benchmarks such as SplitCIFAR-10 have demonstrated ResNet18's strong performance in class-incremental OCL scenarios. ResNet18 also performs well in transfer learning, efficiently adapting to new data streams by leveraging knowledge from previous tasks, making it a preferred choice for CL and OCL applications \cite{NIPS2017_0efbe980}. The study~\cite{lee2024impact} shows that smaller models like ResNet18 or Slim-ResNet18 are better suited for maintaining long-term stability and effectively retaining tasks. Additionally, various modified versions of ResNet18 are employed in OCL approaches, including Reduced, Slim, Lightweight, and Randomly Initialized variants, with the Reduced version being the most widely used.

\textbf{Fully-Connected Networks}, particularly \textbf{Multilayer Perceptrons} (MLPs) \cite{rumelhart1986learning}, have been utilized in 27 approaches, valued for their simplicity and efficiency in adapting to task-specific requirements. MLPs allow models to retain prior knowledge by transforming shared representations while keeping computational and memory demands low~\cite{pham2021contextual}. They are especially effective for real-time data processing tasks, such as sensor data handling, where quick adaptation to new tasks or patterns is crucial~\cite{schiemer2023online}. However, MLPs often face challenges in capturing complex spatial or temporal relationships, necessitating additional mechanisms to effectively process more dynamic or sequential data~\cite{pham2021contextual, schiemer2023online}.

\textbf{Autoencoders}~\cite{hinton2006reducing} are seldom used, appearing in only 8 approaches. They are employed to efficiently compress and replay data features, helping to reduce memory usage and mitigate catastrophic forgetting \cite{wang2021acae}. For instance, \textbf{ACAE-REMIND}~\cite{wang2021acae} employs feature replay at intermediate network layers, enhancing training flexibility, while \textbf{MERLIN}~\cite{kj2020meta} uses Variational Autoencoders (VAEs) to model task-specific parameter distributions, improving adaptability across tasks. Despite these advantages, challenges remain, such as the complexity of implementation and suboptimal performance in deeper-layer models like \textbf{ACAE-REMIND}.

There were only three studies that used \textbf{Transformers}~\cite{vaswani2017attention} as the main architecture in their approaches. One notable approach is the \textbf{Contrastive Vision Transformer (CVT)}~\cite{10.1007/978-3-031-20044-1_36}, which employs a transformer architecture specifically optimized for the OCL setting. CVT introduces an external attention mechanism to encode past task information while minimizing parameter usage. It uses class-specific learnable vectors, called persistent attention vectors, to retain knowledge of previous tasks. Additionally, it uses stacked transformer blocks to capture global dependencies. By combining these features with a dual-classifier structure, CVT achieves a balance between learning new tasks and preserving knowledge of earlier ones.

\subsubsection{Optimization Techniques}

Among the various optimization techniques, \textbf{gradient-based optimizers}\cite{goodfellow2016deep} were the most widely applied in OCL literature. Among these, \textbf{Stochastic Gradient Descent (SGD)}\cite{bottou2010large} stands out, being used in 36 approaches~\cite{cai2022improving,zou2022efficient}. SGD is effective for handling non-stationary data, enabling continuous adaptation as new information becomes available~\cite{cai2022improving}. Its faster convergence makes it well-suited for real-time applications, while its ability to adapt learning rates ensures reliable convergence when tracking shifting data~\cite{cai2022improving,guo2022adaptive}. SGD’s computational efficiency, particularly its avoidance of second-order derivatives, further establishes it as a practical choice for resource-constrained environments~\cite{zou2022efficient,cai2022improving}.

In addition to SGD, other gradient-based optimizers have been employed in specific approaches. These include \textbf{Adam}\cite{kingma2014adam}, \textbf{Lookahead}\cite{zhang2019lookahead}, \textbf{Steepest Gradient Descent}\cite{wright2006numerical}, \textbf{Proximal Gradient Descent (PGD)}\cite{parikh2014proximal}, and \textbf{bilevel optimization}~\cite{colson2007overview}. These alternatives offer additional flexibility and specialized advantages for various OCL scenarios.

\subsubsection{Loss and Activation Functions}

Among the different types of loss functions, \textbf{Cross Entropy-based}~\cite{rubinstein2004cross} losses are the most commonly used, appearing in 22 approaches. Additionally, \textbf{Mean Square Error (MSE)-based} loss functions~\cite{bishop2006pattern} are employed in 4 approaches. Choosing the appropriate loss function plays a critical role in addressing catastrophic forgetting, as highlighted in several studies~\cite{Delange_2021,goodfellow2015empiricalinvestigationcatastrophicforgetting}.

Modern activation functions, such as \textbf{GELU}~\cite{hendrycks2016gelu}, have the advantage of improving gradient flow\cite{harun2023siesta}. However, they have been utilized in only two studies so far. Despite these advancements, many OCL approaches continue to rely on traditional activation functions, such as \textbf{ReLU}~\cite{10.5555/3104322.3104425}, which is used in 11 approaches, and \textbf{softmax}~\cite{bishop2006pattern}, which is used in 9 approaches. ReLU and softmax thus remain the standard choices in the field.

\subsubsection{Regularization Techniques}
Regularization techniques are integral to regularization-based approaches and often complement other strategies.  One well-known example is Knowledge Distillation (KD)~\cite{hinton2015distillingknowledgeneuralnetwork}, which is commonly used as a backbone in many methods. This technique ensures that older information is not lost during updates~\cite{HAN2022104966,fini2020online}. Moreover, KD reduces memory usage by storing predictions instead of raw data, making it especially useful in memory-constrained environments~\cite{fini2020online}.

\subsubsection{Replay and Memory Buffer Techniques}

\textbf{Experience Replay (ER)}~\cite{lin1992self, rolnick2019experiencereplaycontinuallearning} and its related techniques are among the most widely adopted methods, appearing in 20 approaches. These methods are valued for their ability to mitigate catastrophic forgetting, enhance class discrimination, and facilitate knowledge transfer across tasks~\cite{NEURIPS2020_ca4b5656, pham2021contextual, Guo_2023_CVPR}.

In addition, 21 approaches incorporate memory buffer management techniques, such as \textbf{Episodic Memory}~\cite{tulving1972episodic} and \textbf{replay buffers}, to effectively retain past experiences. Replay-based methods—including \textbf{pure replay}, \textbf{mixed replay}, \textbf{supervised contrastive replay}, and \textbf{self-replay}—offer unique advantages in improving learning efficiency and robustness in OCL scenarios~\cite{lopezpaz2022gradientepisodicmemorycontinual, han2022selecting, mai2021supervised}.

To further enhance memory efficiency, several approaches utilize memory sampling and buffer strategies like \textbf{Reservoir Sampling}~\cite{vitter1985random} and the \textbf{ring buffer}. Reservoir Sampling is a randomized algorithm designed to maintain a uniform sample of size \(k\) from a potentially infinite or unknown stream of data, ensuring equal probability for every element to be included in the sample. On the other hand, a ring buffer (also known as a circular buffer) is a fixed-size data structure that overwrites the oldest elements when new data is added after the buffer is full, making it an ideal choice for real-time systems or applications requiring consistent memory usage.

\subsubsection{Data Augmentation Techniques}

Data augmentation techniques are useful in memory-constrained environments, as they help prevent feature bias and promote comprehensive learning across tasks~\cite{fini2020online, caccia2021reducing, pmlr-v162-guo22g}. Common methods, such as image flips and random grayscale adjustments, have been used in four studies. More advanced techniques, like CutMix~\cite{shorten2019survey, yun2019cutmix}, improve model robustness by combining and modifying image segments to generate diverse and novel training samples.

\subsubsection{Classification and Clustering Methods}
In OCL approaches, various classifiers have been utilized, with the \textbf{Nearest Class Mean (NCM)}\cite{mensink2013distance} classifier being a common choice, appearing in 9 approaches. NCM is effective at avoiding task-recency bias and seamlessly integrating new classes without requiring architectural modifications\cite{mai2021supervised}. Moreover, techniques such as \textbf{Supervised Contrastive Replay} enhance NCM by clustering class embeddings more effectively, further mitigating catastrophic forgetting~\cite{mai2021supervised}. Another widely used method is \textbf{K-Means}~\cite{bishop2006pattern}, a clustering algorithm that partitions data into distinct clusters, offering flexibility and simplicity in organizing data representations. Similarly, \textbf{Softmax classifiers} are frequently employed for their ability to assign probabilistic outputs to classes, making them effective for multi-class problems in OCL. These classifiers are often used in combination with replay buffers or distillation techniques to balance learning across tasks. Additionally, \textbf{Linear Classifiers}, known for their computational efficiency and simplicity, are integrated into OCL frameworks to directly map learned features to class labels. Each of these classifiers is adapted in OCL models to address challenges such as task interference and catastrophic forgetting while maintaining a balance between accuracy and efficiency.

\subsubsection{Libraries and Frameworks}

Among the libraries and tools examined, \textbf{PyTorch}~\cite{paszke2019pytorch} emerges as the most widely used, appearing in 12 studies. Its popularity stems from its flexibility and widespread adoption within the deep learning community. PyTorch provides powerful features such as dynamic computation graphs, extensive support for neural networks, and seamless GPU acceleration, making it an ideal choice for developing and deploying OCL models.

Another notable library is \textbf{Avalanche}~\cite{lomonaco2021avalanche}, built on top of PyTorch and specifically designed for CL. Avalanche offers a comprehensive framework that simplifies experimentation with various CL techniques, including experience replay, regularization, and architecture approaches. Though slightly less common than PyTorch, Avalanche's specialized focus is invaluable for OCL research, with significant potential for improvement.

\subsection{Features}\label{Features}

In this section, we discuss the most significant features identified in the OCL literature. A feature is defined as a setting, problem, or characteristic that distinguishes different approaches. We identified approximately 48 frequently occurring or notable features, which were analyzed and categorized based on their type and purpose across 81 approaches. These features are grouped into seven categories: 1) Settings, 2) Challenges, 3) Data Management, 4) Memory Management, 5) Knowledge Transfer and Adaptation, 6) Optimization, and 7) Efficiency (see Figure~\ref{fig:feature}). The complete list of extracted features and their associated approaches can be found in Appendix~\ref{featureappendix1}.

\begin{figure}[t!]
    \centering
    \includegraphics[width=1\textwidth]{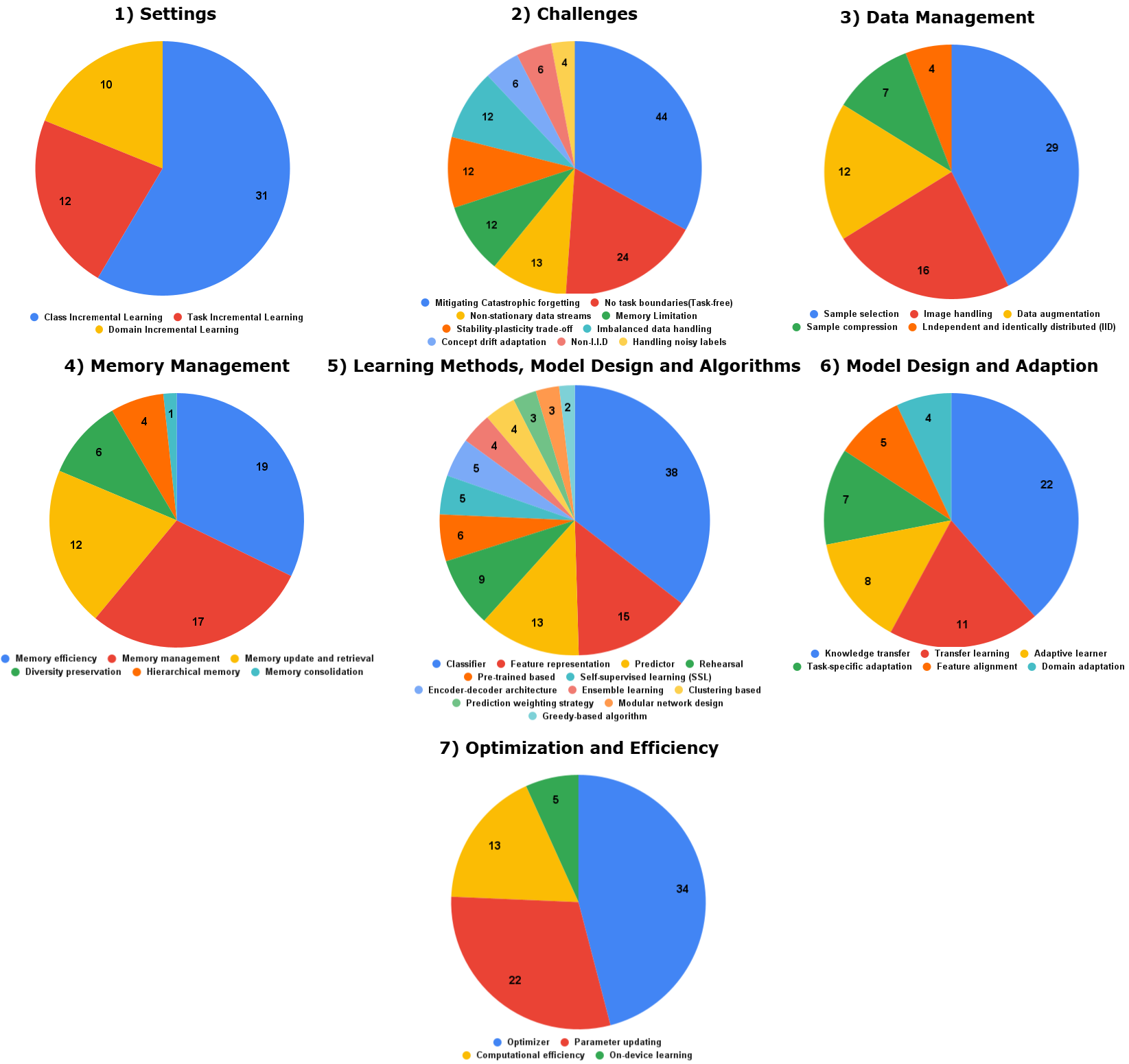}
    \caption{Key features in OCL literature. This figure highlights the most important and prevalent features identified in the literature, with the number of occurrences of each entity across 81 final studies provided for each category.}
    \label{fig:feature}
\end{figure}

\subsubsection{Settings}

A setting refers to a specific scenario or configuration—a set of assumptions and conditions under which a model learns and generates outputs. In OCL, most studies use these three settings: Task Incremental Learning (TIL), Class Incremental Learning (CIL), and Domain Incremental Learning (DIL). Among these three settings, CIL was the most commonly used, appearing in 32 research papers. This popularity can be attributed to the fact that, despite being more challenging than the other settings, CIL better reflects real-world scenarios~\cite{shim2021online}. Additionally, some studies highlighted approaches that operate across multiple settings. For instance, the novel approach \textbf{MERLINE} demonstrated its applicability in both CIL and DIL settings~\cite{kj2020meta}.


\subsubsection{Challenges}

Several key challenges in OCL include (1) mitigating catastrophic forgetting, where models lose knowledge of past tasks while learning new ones, addressed through techniques like replay buffers and parameter isolation; (2) managing the stability-plasticity trade-off, which balances retaining past knowledge (stability) and adapting to new tasks (plasticity) using regularization and architectural adjustments; (3) handling non-stationary data streams, where changing data distributions over time require dynamic memory allocation and online adaptation; and (4) operating without clear task boundaries, addressed by task-agnostic and unsupervised learning methods to seamlessly adapt to unknown transitions. These challenges drive ongoing innovations to enhance OCL robustness and effectiveness. These issues are considered some of the most significant in OCL and have been widely discussed in the literature.

One of the main challenges is mitigating catastrophic forgetting, which involves ensuring that a model retains previously learned knowledge when acquiring new information~\cite{parisi2019continual}. In machine learning, when a model is retrained on new and different data, its parameters are updated based on this new input. However, these updates can cause the model to lose its ability to correctly predict outcomes for the previous data, leading to what is called catastrophic forgetting~\cite{kirkpatrick2017overcoming}. Addressing this issue is crucial in OCL, and it is one of the most frequently mentioned challenges, appearing in 44 studies. Additionally, challenges related to the nature of input data play a major role in OCL literature. These include the lack of clear task boundaries (task-agnostic scenarios), noisy or imbalanced data, overly complex data, non-stationary data streams, and non-i.i.d. data.

\subsubsection{Data Management}

In OCL setup, the input data arrives sequentially, either as individual samples or in small batches, while storage memory is limited~\cite{wiewel2021entropy}. Managing this incoming data efficiently is critical for maintaining the performance of learning approaches, especially when handling imbalanced or non-i.i.d. data or when processing large data volumes without well-defined task boundaries~\cite{wiewel2021entropy, aljundi2019gradient}.

Replay-based strategies must address the challenge of limited memory capacity, which requires removing older stored data to make space for new data. Without replaying past data, models are at risk of catastrophic forgetting~\cite{sangermano2022sample}. To enhance both accuracy and efficiency, the data selected for replay should be balanced and diverse~\cite{bang2022online} while minimizing memory usage~\cite{10444954}. Techniques for sample selection focus on choosing representative samples that maximize diversity and ensure class balance, while sample compression methods help reduce the memory footprint~\cite{10444954}. For example, the \textbf{GSS} approach~\cite{aljundi2019gradient} employs a greedy algorithm for sample selection, whereas the \textbf{OLCGM} approach~\cite{sangermano2022sample} merges samples to achieve compression.

The limited memory size also restricts the number of samples available for replay, increasing the risk of overfitting. To address this, some methods, like \textbf{RAR}~\cite{zhang2022simple}, use data augmentation techniques. However, in scenarios with extreme memory constraints, standard data augmentation may not be practical, as it can significantly increase memory usage~\cite{fini2020online}.

\subsubsection{Memory Management}

Memory management in OCL focuses on achieving memory efficiency by storing only the most essential information while reducing overall memory usage~\cite{gu2024summarizing}. Since OCL systems operate under strict memory constraints, it is important to retain a diverse set of useful information to ensure consistent model performance~\cite{bang2021rainbow}. Efficient memory management allows the system to store, update, and retrieve data effectively, enabling access to relevant past experiences without compromising accuracy over time~\cite{chaudhry2018efficient, gu2024summarizing, bang2021rainbow}. A critical part of memory management is the process of memory updating and retrieval, which is particularly important in Replay-based strategies. This process involves refreshing the memory with a mix of new and previously stored data during each learning cycle. Furthermore, maintaining memory diversity is essential. Retaining a wide range of data samples from different classes, tasks, or domains ensures that the system does not overfit to specific patterns, which helps improve generalization and leads to more robust performance~\cite{aljundi2019gradient}.

\subsubsection{Learning Methods, Model Design and Algorithms}

The ``Learning Methods, Model Design, and Algorithms" category contains a wide range of techniques, architectures, and strategies developed to address the unique applications in OCL. These methods often focus on solving specific tasks, such as classification, prediction, or clustering. For instance, the \textbf{ONLINE COMPRESSION SQM} approach~\cite{caccia2019online} is designed specifically for image classification.

In some cases, these methods also target specific components within an OCL approach. For example, \textbf{OCL-HAR}~\cite{schiemer2023online} employs a classifier to predict incoming data while clustering outliers, which are then stored in memory for replay. To further enhance performance, various learning algorithms have been proposed to tackle tasks like sample selection from incoming data streams~\cite{aljundi2019gradient}. An illustrative example is \textbf{DISCOBER}~\cite{seo2024just}, which uses an ensemble method to diversify generated images for replay, thereby improving the learning process.

\subsubsection{Model Design and Adaption}

This category focuses on the ability to learn from new data while effectively transferring previously acquired knowledge to the model. Efficient knowledge transfer during each training phase is essential to minimize catastrophic forgetting~\cite{pham2020bilevel}. At the same time, adapting the model to new data and optimizing the learning process are equally critical. Many approaches aim to improve at least one of these aspects~\cite{cai2021online}.

For instance, \textbf{Continual-MAML}~\cite{caccia2020online} introduces a mechanism that enables selective forgetting, facilitating faster and more efficient adaptation to new data. On the other hand, \textbf{AMA+MALR}~\cite{cai2022improving} addresses the challenge of retaining information, emphasizing that even under optimal resource conditions, retention remains a significant issue, particularly in large-scale OCL scenarios.

\subsubsection{Optimization and Efficiency}

As new data arrives, the model must update its weights before the next batch comes in, ensuring that it learns new information. However, this process is constrained by time, computational limits, and memory restrictions, making it especially critical~\cite{cai2021online}. Consequently, many research papers focus not only on improving accuracy and mitigating catastrophic forgetting but also on reducing computational costs and optimizing memory usage~\cite{yang2022online, cai2021online, zou2022efficient, harun2023siesta}. For example, the \textbf{SIESTA} approach~\cite{harun2023siesta} highlights real-world applications of OCL, particularly in resource-constrained devices with limited memory and computational resources. It proposes a method that aligns with the conditions of on-device learning~\cite{dhar2021survey}.

Hyperparameter optimization poses a critical challenge during the OCL phase, as hyperparameters must be recalibrated whenever data distributions shift. Selecting appropriate values, such as learning rate and batch size, is complex, as improper choices can significantly hinder the model's future performance~\cite{prabhu2023online}. To address this, the \textbf{RAR} approach~\cite{zhang2022simple} introduces a reinforcement learning-based method that dynamically adjusts hyperparameters, effectively managing the stability-plasticity trade-off in real-time.

\subsection{Quality Attributes}\label{QA}

Quality attributes are criteria, features, or metrics that describe the overall approach or its specific components. They are vital for the success of OCL approaches, as their variation can either facilitate or impede the achievement of OCL objectives. We identify 51 key quality attributes, categorized into five groups based on their nature: 1) Performance Metrics, 2) Efficiency, 3) Model Capacity and Flexibility, 4) Detection and Data Distribution, and 5) Others. A mapping table of approaches and their quality attributes can be found in Appendix Section~\ref{qualityappendix1}.

The \textbf{Performance Metrics} category includes attributes related to accuracy, efficiency, and generalization. The aim is to improve accuracy and generalization while minimizing overfitting and complexity. The \textbf{Efficiency} category focuses on memory usage, optimization, and overall resource consumption, where the goal is to reduce memory usage and improve efficiency. For example, the \textbf{Entropy-based Sample Selection} method~\cite{wiewel2021entropy} addresses memory optimization by avoiding static buffers and using random sample replacement, which helps reduce overfitting.

Attributes concerning the flexibility of methods or components for learning new data are grouped under \textbf{Model Capacity and Flexibility}. The \textbf{Detection and Data Distribution} category focuses on preserving past knowledge, ensuring the retention of previously learned information, and preventing forgetting while acquiring new data from different distributions. Reducing computational and time costs is a goal shared across all categories. For instance, \textbf{MuFAN}~\cite{jung2023new} achieves high stability and strong plasticity by introducing a novel distillation loss and a normalization module.

\subsection{Datasets}\label{Datasets}

\begin{figure*}[t!]
    \centering
    \includegraphics[width=1\textwidth]{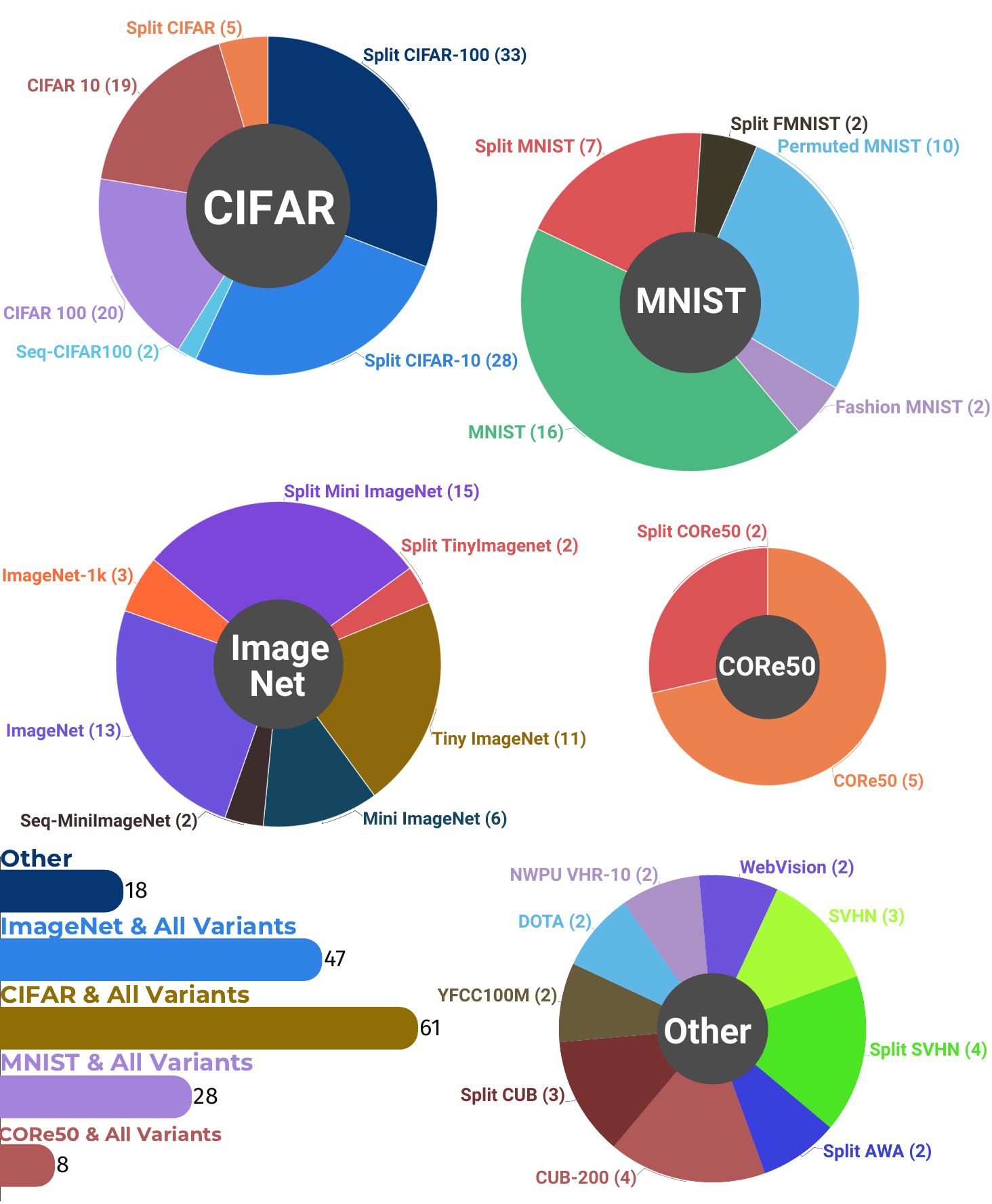}
    \caption{Prominent datasets in OCL literature. This figure categorizes and lists key datasets employed in OCL research, featuring CIFAR, ImageNet, MNIST, and CORe50 variants.}
    \label{fig:dataset}
\end{figure*}

\begin{figure}[t!]
    \centering
    \includegraphics[width=1\linewidth]{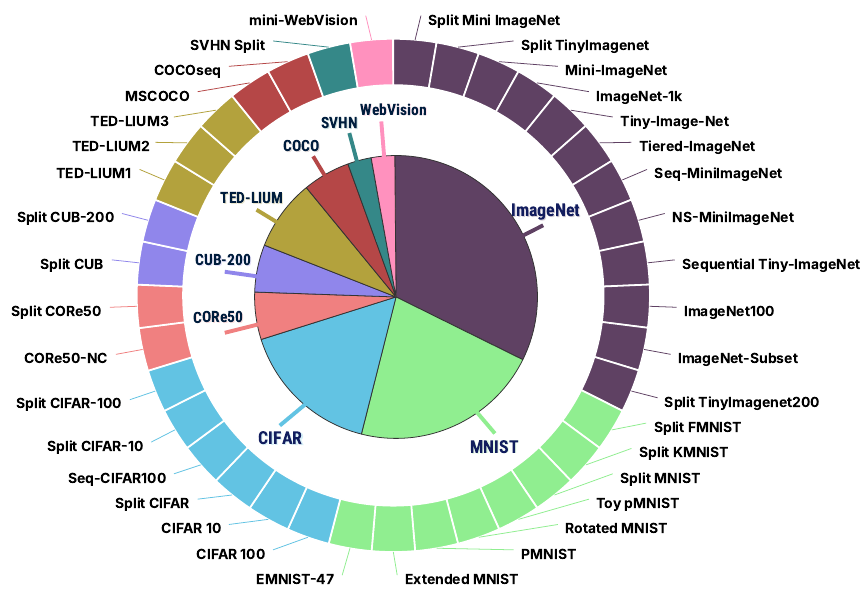}
    \caption{Variants of datasets in OCL literature. This figure visualizes the relationships between major datasets and their corresponding variants. Dataset names are arranged in the inner circle, with their variants positioned in the outer circle. A consistent color scheme links each dataset to its variants.}
    \label{fig:dataset-variants}
\end{figure}

The datasets in OCL literature originate from diverse fields, each designed to address real-world challenges (see Figure~\ref{fig:dataset}). Figure~\ref{fig:dataset-variants} illustrates the key variants of these datasets, which we categorize into four main types: (1) Image Classification, (2) Object Detection and Segmentation, (3) Multimodal Vision-Language, and (4) Activity Recognition and Tracking. Details on dataset usage in each study are found in Appendix~\ref{datasetappendix}.

\subsubsection{Image Classification}

One of the most common datasets used in OCL research is the \textbf{CIFAR} datasets~\cite{krizhevsky2009learning}, which appear in 61 studies. These datasets, created by the Canadian Institute for Advanced Research, contain labeled images that are often used to test computer vision models. The \textbf{CIFAR-10} and \textbf{CIFAR-100} datasets have images from 10 and 100 classes, respectively. Variants such as \textbf{Split CIFAR-10}~\cite{zenke2017continual} and \textbf{Split CIFAR-100} divide the classes into smaller groups that are presented in sequence, which makes them very useful for OCL studies.

Other well-known datasets in this area are \textbf{MNIST}~\cite{lecun1998gradient} and \textbf{ImageNet}~\cite{deng2009imagenet}, which are used in 28 and 47 studies, respectively. MNIST, a famous dataset with handwritten digits, has several variations. For example, \textbf{Permuted MNIST (pMNIST)}\cite{kirkpatrick2017overcoming} randomly shuffles the pixels of the images to create new tasks, and \textbf{Split MNIST}\cite{zenke2017continual} splits the digit classes into separate tasks. \textbf{Split Fashion MNIST (Split FMNIST)}~\cite{xiao2017fashion} follows the same idea but uses images of clothing items instead.

\textbf{ImageNet}, known for its large and diverse set of images, has also been adjusted for simpler tasks. \textbf{Mini ImageNet}~\cite{vinyals2016matching} and \textbf{Tiny ImageNet} reduce the image resolution and the number of classes, making them more manageable for specific OCL scenarios. There are also other datasets like \textbf{Tiered-ImageNet}\footnote{\href{https://github.com/yaoyao-liu/tiered-imagenet-tools}{https://github.com/yaoyao-liu/tiered-imagenet-tools}}, \textbf{ImageNet100}\footnote{\href{https://www.kaggle.com/datasets/ambityga/imagenet100}{https://www.kaggle.com/datasets/ambityga/imagenet100}}, and \textbf{Places365-Standard}~\cite{zhou2017places}, which add more classes and complexity, helping to test how well models scale and perform under more difficult conditions.

\subsubsection{Object Detection and Segmentation}

Several datasets are commonly used in OCL research for object detection and segmentation tasks. These datasets help test models in areas such as autonomous driving and remote sensing. Some well-known datasets include \textbf{MSCOCO}~\cite{10.1007/978-3-319-10602-1_48}, \textbf{KITTI}~\cite{geiger2013vision}, \textbf{SemKITTI-DVPS}~\cite{qiao2021vip}, \textbf{Cityscapes}~\cite{cordts2016cityscapes}, \textbf{Oxford RobotCar}~\cite{maddern20171}, \textbf{DOTA}~\cite{xia2018dota}, \textbf{NWPU VHR-10}\cite{NWPU}, \textbf{RSOD}~\cite{10292941}, and \textbf{DIOR}~\cite{li2020object}. These datasets are used in different applications, such as testing self-driving cars (e.g., KITTI and Cityscapes) or analyzing aerial images (e.g., DOTA and DIOR). They pose different challenges, such as varying weather conditions, camera types, and object sizes, which are useful for testing OCL models in real-world environments.

\subsubsection{Multimodal Vision-Language}

Another type of dataset was multimodal vision-language, which combines both visual and textual data. These datasets require models to process and integrate information from different modalities effectively. Key datasets in this category include \textbf{YFCC100M}~\cite{thomee2016yfcc100m}, a large collection of 100 million images and videos under Creative Commons, \textbf{NUS-WIDE}~\cite{chua2009nus}, which has images annotated with 81 concepts for multi-label classification and retrieval, and \textbf{WebVision}~\cite{li2017webvision}, which is a benchmark for learning visual representations from noisy web data. Additionally, the \textbf{COCOseq}~\cite{pham2022continualnormalizationrethinkingbatch} dataset is a continual learning version of the popular COCO dataset and is used for sequential tasks involving both vision and language. These datasets are used for tasks like image captioning, cross-modal retrieval, and aligning visual and textual data.

\subsubsection{Activity Recognition and Tracking}

OCL applications in activity recognition and tracking require models that can identify human activities or track objects over time in real-time situations. Datasets like \textbf{WISDM}~\cite{schiemer2023online}, which includes data from smartphones and smartwatches for activity recognition, \textbf{HAPT}~\cite{schiemer2023online}, which focuses on human activity recognition, \textbf{PAMAP2}~\cite{6246152}, used for physical activity monitoring, and \textbf{PETS}~\cite{8014998}, which is used for evaluating tracking systems in surveillance, are commonly used in these areas. These datasets include data from wearable sensors and video cameras, which present challenges like recognizing complex activity patterns and ensuring accurate tracking in constantly changing environments.

\subsection{Evaluation metrics}\label{metrics}

While measuring the accuracy of proposed approaches is crucial, it is equally important to assess how well these methods mitigate or exacerbate the existing challenges in this domain. Beyond average accuracy, which remains a widely used metric, many approaches incorporate additional evaluations to monitor the extent of forgetting and the ability to learn new tasks effectively. In this section, we review some of the most common and significant evaluation metrics.

\textbf{Average Accuracy}: represent the accuracy of the model on the test set of task \textit{j} after training on task \textit{i}, where \textit{i} and \textit{j} refer to tasks in the sequence. The average accuracy by the end of training on task \textit{T} can be defined as: 
\[A_T = \frac{1}{T}\sum_{j=1}^{T} a_{T,j}\]
This metric reflects the model's overall performance after learning the entire task sequence, capturing its ability to generalize across all tasks. A higher \(A_T\) indicates better retention and learning but does not account for forgetting or other specific behaviors of continual learning algorithms\cite{Chaudhry_2018,chaudhry2019tiny,mai2021supervised}.

\textbf{Forgetting Measure}: Forgetting for a task \textit{j} is defined as the difference between the maximum accuracy achieved on that task before learning task \textit{k} and the accuracy on the same task after training on task \textit{k}. This quantifies how much the model has forgotten about previous tasks as new tasks are learned. Formally, forgetting for task \textit{j} after training up to task \textit{k} is computed as: 
\[
f_{k,j} = \max_{l \in \{1, \dots, k-1\}} a_{l,j} - a_{k,j}, \quad \forall j < k
\]
The \textbf{Average Forgetting Measure} at task \textit{T} is then given by: 
\[
F_T = \frac{1}{T-1} \sum_{j=1}^{T-1} f_{T,j}
\]
A lower value of \(F_T\) indicates less forgetting, meaning the model retains more knowledge of previously learned tasks\cite{Chaudhry_2018,chaudhry2019tiny,pmlr-v162-guo22g}.

\textbf{Learning Accuracy:} Learning Accuracy measures the model's performance on each task immediately after it has been trained on that task. This metric reflects the model's ability to acquire new knowledge. Formally, the Learning Accuracy at task \( T \) is defined as:

\[
LA = \frac{1}{T} \sum_{i=1}^{T} a_{i,i}
\]

A higher \( LA \) indicates better performance in learning new tasks, with the task identifiers used only for evaluating this metric\cite{pham2022continualnormalizationrethinkingbatch,pham2021contextual,Han_2023}.

\textbf{Mean Average Precision (mAP):} The mean average precision (mAP) is a widely adopted evaluation metric used to assess the performance of multi-class object detection algorithms. It calculates the average precision (AP) across all classes, measuring how well a model identifies the correct bounding boxes for objects. AP is driven by recall and precision, and it is generally calculated separately for each class. If the Intersection over Union (IoU) between the predicted bounding box and the ground truth is higher than 0.5, the bounding box is considered correct. Precision and recall are defined as:
\[{Recall} = \frac{TP}{TP + FN}\]\[{Precision} = \frac{TP}{TP + FP}\]
Where \(TP\), \(FP\), and \(FN\) represent the true positives, false positives, and false negatives, respectively. A higher mAP score indicates better model performance across all classes\cite{CHEN2023105549,jiang2021multi,kwon2022toward}.

\textbf{Backward Transfer (BWT):} Backward transfer (BWT) measures the influence that learning a new task \( t \) has on the performance of previously learned tasks \( k \) where \( k < t \). Positive backward transfer occurs when learning a task improves performance on preceding tasks, while negative backward transfer indicates that learning a task reduces performance on earlier tasks. Large negative backward transfer is often referred to as catastrophic forgetting. BWT is formally defined as:

\[
BWT = \frac{1}{T - 1} \sum_{i=1}^{T-1} (a_{T,i} - a_{i,i})
\]

A higher value of BWT suggests better retention of previous knowledge and less forgetting. If two models have similar accuracy, the model with the higher BWT is considered to have better knowledge retention and transferability\cite{lopezpaz2022gradientepisodicmemorycontinual, HAN2022104966, Chaudhry_2018}.

\section{Discussion}\label{discussion}

\subsection{Address Research Questions}

In this section, we revisit the research questions outlined at the start of the study and provide insights derived from the analysis and results discussed in Section~\ref{analysis}. 

\paragraph{RQ1 - Problem Setup in OCL}

OCL involves processing sequential data streams in real-time, requiring models to learn continuously without revisiting past data. To operate in an OCL setup, three key conditions must typically be met: (1) data is fed to the model in real-time and in small batches, (2) tasks at different time steps have disjoint label spaces, and (3) each task's data is trained in a single pass, with no opportunity to revisit earlier data. Our analysis highlights the unique challenges inherent in OCL, such as non-stationary data distributions, imbalanced data streams, and the absence of task boundaries in task-free setups. These challenges necessitate efficient methods to balance stability and adaptability, ensuring that models prevent catastrophic forgetting while incorporating new knowledge. OCL settings, including task-incremental, class-incremental, domain-incremental, and task-free learning, provide flexibility in addressing these challenges. Each variant varies in complexity, with task-free learning being the most challenging. This is because it lacks explicit task identifiers, requiring models to detect changes in data distribution autonomously.

\paragraph{RQ2 - Approaches and Methods}

OCL methods tackle specific challenges using three main strategies: replay-based, architecture-based, and regularization-based approaches. Replay-based methods, found in 62 of the 81 analyzed approaches, rely on memory buffers or generative models to revisit past data or generate synthetic examples. Experience replay models help balance memory usage with model performance, but they require careful tuning to manage memory limitations. Architecture-based approaches adjust the model structure by expanding capacity or isolating task-specific parameters. Techniques like \textbf{Progressive Neural Networks} showcase the effectiveness of parameter isolation, though scalability becomes an issue as the number of tasks increases. Regularization-based methods, while memory-efficient, often struggle with long task sequences. These methods constrain weight updates to preserve previous knowledge, utilizing techniques such as knowledge distillation and elastic weight consolidation.

\paragraph{RQ3 - Common Data Types, Datasets, and Benchmarks}

The variety of datasets and benchmarks in OCL research underscores its broad applicability across diverse domains. Our analysis identified 83 distinct datasets used in this field. Based on the analysis of 81 selected studies, these datasets are grouped into four main categories: Image Classification, Object Detection and Segmentation, Multimodal Vision-Language, and Activity Recognition and Tracking. Image classification datasets, such as CIFAR-10, CIFAR-100, MNIST, and ImageNet, are widely used and often appear in split or permuted versions to simulate continual learning scenarios. For object detection and segmentation, datasets like MSCOCO, KITTI, Cityscapes, and DOTA are commonly employed to evaluate models in complex environments, including autonomous driving and aerial imagery, where challenges like varying conditions and object diversity arise. Multimodal vision-language datasets, including YFCC100M, NUS-WIDE, and COCOseq, combine visual and textual data, supporting tasks such as image captioning and cross-modal retrieval. Lastly, activity recognition and tracking datasets, such as WISDM, HAPT, PAMAP2, and PETS, focus on identifying human activities and tracking objects over time, often involving data from wearable sensors and video cameras to handle real-time dynamic scenarios. Together, these datasets provide a broad basis for evaluating OCL models across different tasks and environments.

\paragraph{RQ4 - Components and Features of Proposed Approaches}

OCL methods integrate various components and features to achieve their specific objectives. Common model architectures, such as ResNet variants—particularly ResNet18—strike a balance between computational efficiency and accuracy. ResNet architectures, with their residual connections, help address vanishing gradient issues, ensuring stable learning across tasks. Optimization techniques like stochastic gradient descent (SGD) and its variants allow effective adaptation to non-stationary data. Memory management techniques, including reservoir sampling and episodic memory buffers, are crucial for replay-based methods, providing efficient ways to retain essential data points. Key features and components identified include mechanisms to combat catastrophic forgetting, such as knowledge distillation and data augmentation techniques like CutMix, which improve the diversity and robustness of training data. Together, these components and features form the foundation of OCL methods, enabling models to function effectively in dynamic, resource-constrained environments.

\paragraph{RQ5 - Tools and Libraries in OCL Research}

The widespread use of tools like PyTorch and Avalanche highlights their essential role in advancing OCL research. PyTorch’s flexibility and comprehensive ecosystem make it an ideal platform for implementing various OCL strategies, ranging from replay-based methods to architectural adaptations. Avalanche, a specialized library built on PyTorch, simplifies experimentation with CL approaches by providing modular implementations of replay buffers, regularization techniques, and architectural strategies. Their role goes beyond implementation, offering built-in evaluation metrics and datasets that support high-quality benchmarking of OCL methods. The adoption of these tools emphasizes the growing need for robust frameworks capable of supporting the complex requirements of OCL research.

\subsection{Practical Implications}

In this subsection, we highlight the key practical implications of our work for the research community. We provided a comprehensive list of extracted entities, including 81 approaches, 127 components, 48 features, 60 quality attributes, and 83 datasets, each accompanied by concise definitions. We hope that this resource enhances researchers' understanding of the various components and features utilized in OCL approaches.

Furthermore, the tables~\ref{componentsappendix1} and~\ref{componentsappendix2} offer valuable insights into how different components are combined within specific approaches. This enables researchers to evaluate the compatibility of components in addressing challenges and explore new solutions through these combinations.

In addition to Newman's guidelines, we aim to share the steps taken in our SLR process, providing a clear and transparent methodology for other researchers. By documenting each stage—from the initial search strategy to the final selection criteria—we hope to establish good practices for conducting SLRs in various fields. We believe adopting an SLR approach will encourage a thorough exploration of existing literature while helping researchers avoid potential biases and gaps in their analyses.

\subsection{Limitations of the Frequency-Based Approach}

Our work employs a frequency-based method to extract and analyze data, offering insights into how often certain entities are discussed in the literature and identifying prevailing trends. 
While this approach offers a broad overview, it also presents notable limitations that require further discussion. First, this method lacks the depth required to delve into the theoretical underpinnings or algorithmic details of different approaches. Unlike studies such as \cite{parisi2020online}, which explore these aspects comprehensively, our analysis does not illuminate the foundational principles that explain the efficacy or application of specific methods. Second, our frequency-based approach falls short of capturing the nuanced empirical comparisons frequently presented in the literature. For instance, studies like \cite{mai2022online} conduct detailed evaluations to assess the strengths and weaknesses of various methods. In contrast, our method overlooks these in-depth comparative analyses, which are crucial for understanding the practical implications and trade-offs among different approaches.

\subsection{Suggestions}

Based on our review of the most significant research papers in this field, we offer several suggestions to researchers. We hope these recommendations will enhance the quality of research and encourage a more collaborative and cohesive community.

First and foremost, \textbf{it is essential to clearly explain the rationale for using different components in approaches}. One of our initial objectives was to create a mapping between components and their corresponding features. For example, if ResNet18 is used as a component in a particular approach, we aimed to connect it to the specific features that justify its inclusion. However, many of the papers we reviewed failed to explain their component choices adequately, which led to significant sparsity in the mapping table. As a result, we decided not to include this mapping table due to the lack of clarity in the reviewed studies.

Another notable observation is the lack of clear explanations regarding the metrics used in some research papers. We found that some metrics in the field are functionally identical but are often named differently. While this is somewhat understandable, given that OCL is a relatively new and evolving field, it still leads to confusion and inconsistency. To prevent duplication of terminology and ensure clarity, we recommend that researchers provide precise mathematical definitions for the metrics they employ. For example,~\cite{pham2021contextual} adapted the \textbf{Average Accuracy (ACC)} metric from~\cite{lopezpaz2022gradientepisodicmemorycontinual}, while~\cite{bonicelli2023effectiveness} defined \textbf{Final Average Accuracy} using the same formula. Similarly, \textbf{Forgetting Measure}~\cite{pham2020bilevel} and \textbf{Average Forgetting}~\cite{shim2021online} were other examples of metrics with overlapping definitions and formulas.

\textbf{Reproducibility is fundamental to science}, as it ensures that research findings can be validated and built upon by others in the community. Sharing code, benchmarks, and methodologies associated with state-of-the-art approaches is essential for this validation process and can help propel the field forward by fostering collaboration and innovation. Unfortunately, our results showed that only 20\% of the papers we examined provided links to their code repositories. This lack of openness is concerning because it impedes other researchers' ability to replicate results and contributes to a fragmented knowledge base. To promote progress and improve the reliability of scientific contributions, we strongly encourage researchers to share their code and resources publicly. This openness can significantly advance the growth and success of the entire research community.

\subsection{Future Directions}

Although significant progress has been made in OCL, many challenges remain unresolved. This section outlines key areas that could benefit from deeper investigation and development.

\subsubsection{Catastrophic Forgetting and the Stability-Plasticity Trade-off}

One of the core challenges in OCL is achieving the right balance between \textbf{stability}—the ability to retain previously learned knowledge—and \textbf{plasticity}—the capacity to adapt to new tasks. Several methods have been proposed to address this trade-off specifically. For instance, \textbf{MuFAN}~\cite{jung2023new} employs regularization techniques to preserve previously learned knowledge by penalizing changes to important model parameters. Similarly, \textbf{TRAF}~\cite{kong2023trust} introduces trust-based adjustments, ensuring that updates for new tasks do not significantly interfere with established knowledge. Despite these advancements, challenges remain, especially in dynamic, multi-task environments where task sequences are unpredictable and data distributions shift over time.

Hybrid approaches have shown promise in addressing these limitations. For example, combining regularization with techniques like \textbf{equivariant regularization}, as demonstrated by \textbf{CLER}~\cite{bonicelli2023effectiveness}, enhances the robustness of models against transformations in task structure. A practical scenario where this proves beneficial is in robotics, where tasks may involve variations in spatial configurations or object dynamics.

Exploring such methods further in unstructured and evolving task distributions, such as real-world sensor data streams or user behavior modeling, could lead to significant improvements. Moreover, leveraging \textbf{neural architecture search (NAS)} to automatically design optimal network structures tailored for complex, multi-task environments holds significant potential. For example, NAS could identify architectures that allocate modular components to different tasks, enhancing both stability and plasticity.

Finally, integrating OCL strategies into \textbf{federated learning systems} presents an exciting avenue. For instance, in a federated learning setup for healthcare, where distributed clients (hospitals) encounter evolving patient data, continual learning can help adapt to new patterns (e.g., disease outbreaks) while retaining prior knowledge. This synergy could provide valuable insights for developing robust distributed OCL frameworks, paving the way for efficient and scalable solutions.

\subsubsection{Memory Efficiency and Scalability}

Efficient memory management remains a critical issue in OCL, especially when systems operate under limited resources. Approaches like \textbf{SSD}~\cite{gu2024summarizing} have made strides in memory replay by summarizing key information for storage, yet challenges remain in optimizing memory architectures to balance performance with resource constraints. For example, in resource-limited environments such as mobile devices or edge computing, effective memory management is essential to ensure real-time learning without sacrificing accuracy.

Generative models, such as those used in the \textbf{CONAN} framework, offer an innovative solution by synthesizing data to reduce memory usage~\cite{seo2024just}. These models eliminate the need for explicitly storing raw data, instead reconstructing representative samples as needed. However, their real-time applicability still requires further refinement, mainly in scenarios involving high-frequency data streams, such as financial market predictions or IoT sensor networks.

As real-world applications often deal with large and complex datasets, scalability becomes an even more pressing challenge. For instance, models must handle datasets that grow dynamically, such as user activity logs on social media platforms or geospatial data for climate modeling. Solutions like \textbf{SCALE}~\cite{dam2022scalable} have demonstrated promise by enabling scalable memory replay mechanisms, but extending these solutions to intricate tasks like time-series forecasting (e.g., predicting stock prices) and geospatial analysis (e.g., mapping flood risks) is an ongoing research problem.

Future work could focus on hybrid memory architectures that combine sparse retrieval techniques with generative replay methods. For example, a system might store key indices of past data for quick access while relying on generative models to reconstruct additional data on demand. This approach has the potential to improve both memory use and computational efficiency. Furthermore, incorporating adaptive memory allocation strategies, where memory usage dynamically adjusts based on task complexity and priority, could unlock the potential for real-time applications in large-scale environments, such as autonomous vehicle fleets or distributed monitoring systems.

\subsubsection{Generative Replay and Data Synthesis}

Generative replay is an effective strategy for reducing dependency on stored samples by synthesizing data and helping mitigate catastrophic forgetting. Techniques like \textbf{PCR}, which integrates generative models with contrastive learning, have demonstrated improvements in memory efficiency~\cite{lin2023pcr}. For example, in image classification tasks, generative replay can synthesize diverse representations of previously seen classes, reducing the need for storing large datasets. However, ensuring the quality and diversity of the generated data remains a significant challenge. Low-quality synthetic data could introduce noise or bias, potentially degrading model performance, particularly in critical applications such as medical diagnosis or autonomous navigation.

To address these issues, research could focus on enhancing methods to generate more representative and diverse data for various tasks. For instance, using techniques like \textbf{variational autoencoders (VAEs)} or \textbf{diffusion models} could improve the quality of synthetic samples by capturing more complex data distributions. Additionally, introducing \textbf{feedback loops} within generative models could enable the system to learn from its own outputs and continuously adapt to evolving data streams, making it more suitable for real-time applications like fraud detection or dynamic pricing systems.

Combining adversarial networks with generative replay presents another promising direction. By integrating \textbf{Generative Adversarial Networks (GANs)}, the system could refine the fidelity of synthetic data through adversarial training, ensuring it better reflects real-world distributions. For instance, in tasks involving weather forecasting, adversarial training could ensure synthetic data mimics real atmospheric patterns, enhancing model robustness.

Extending these techniques to handle multi-modal data, such as combining visual and audio inputs, would significantly expand the applicability of generative replay. For example, in speech recognition systems, generating synchronized audio-visual samples could improve performance under noisy conditions. Similarly, multi-modal generative replay could benefit robotics, where systems must process and integrate data from cameras, microphones, and tactile sensors simultaneously. Such advancements would enable generative replay to support a broader range of OCL tasks, paving the way for more versatile and effective systems.

\subsubsection{Multimodal and Non-Visual Tasks}

Most OCL literature has predominantly focused on visual data, leaving non-visual and multimodal tasks underexplored. Expanding frameworks like \textbf{SIESTA} to incorporate additional data types, such as audio, text, and time-series, could significantly broaden the scope of OCL applications~\cite{harun2023siesta}. For instance, in \textbf{speech recognition}, an OCL system could continually adapt to new accents or languages by integrating audio data with linguistic context. Similarly, in \textbf{healthcare}, multimodal learning could combine time-series data from wearable devices with textual data from patient records, enabling more accurate predictions of health outcomes.

Integrating different modalities could also help overcome scenarios where visual data is unavailable or insufficient. For example, in environmental monitoring, combining audio data from sensors that detect animal calls with time-series data such as temperature or humidity readings could provide a more comprehensive understanding of ecological changes. This multimodal approach enhances the system's adaptability, enabling it to adapt to diverse applications beyond traditional computer vision tasks.

A promising direction for advancing multimodal OCL is to explore self-supervised learning, where unsupervised pretraining across modalities could address challenges like data scarcity. For example, a self-supervised framework could learn shared representations of audio, video, and text data, thereby improving the system's generalization capabilities for tasks such as multimedia retrieval.

Incorporating temporal and contextual dependencies into multimodal frameworks could further enhance the system's ability to process sequential data. For instance, in the context of autonomous driving, OCL systems could process real-time audio cues such as honking, visual data like traffic signs, and time-series data, including vehicle speed, to make more accurate decisions. Similarly, in the area of language modeling, integrating temporal dependencies could help the system adapt to evolving language trends or context-specific phrases.

Expanding OCL research to address multimodal and non-visual tasks has the potential to unlock new applications and improve the robustness and adaptability of continual learning systems in real-world scenarios.

\subsubsection{Handling Label Noise and Task Uncertainty}

In dynamic environments, where data can be noisy and tasks evolve over time, handling label noise and task uncertainty is a critical challenge for OCL models. Techniques like selective sampling have shown promise in prioritizing high-quality data during training~\cite{harun2023siesta}. For example, in speech recognition, selective sampling can focus on clearer audio samples to improve model robustness. However, existing methods often struggle with detecting and mitigating noise in large, complex datasets, such as social media text streams or real-time sensor data. More research is needed to develop advanced noise detection and reduction methods, such as using ensemble models or contrastive learning to identify inconsistencies in noisy labels.

Task uncertainty—where task boundaries are unclear or tasks change gradually—adds further complexity to the learning process. For instance, in personalized recommendation systems, user preferences may shift gradually over time, making it challenging to define clear task boundaries. Combining \textbf{self-supervised learning} with probabilistic methods could enhance the adaptability of OCL systems in such uncertain conditions. For example, self-supervised models could learn robust representations of evolving user behavior patterns, while probabilistic techniques model the uncertainty in task transitions.

Future research could explore \textbf{unsupervised task discovery} to automatically identify new tasks without requiring explicit labels or boundaries. For instance, in environmental monitoring, an OCL system could discover emerging patterns in data, such as changes in weather conditions or new species activity, without prior task definitions. Similarly, \textbf{multi-task learning} could be employed to train systems on overlapping tasks, allowing them to learn shared representations that are less sensitive to noisy labels. Additionally, integrating uncertainty estimation with \textbf{task-aware reinforcement learning} could help OCL systems better manage task ambiguity and adapt in real-time. For instance, in robotics, uncertainty-aware models could dynamically adjust their actions based on unclear task objectives or noisy sensor inputs. Developing such techniques would significantly enhance the robustness and flexibility of OCL systems in dynamic and uncertain environments.

\section{Conclusion}
This study provides a detailed analysis of OCL approaches, categorizing common strategies and highlighting key trends, challenges, and research gaps. Reviewing 81 different approaches in OCL highlights the critical importance of replay-based strategies in addressing catastrophic forgetting. We also highlight the growing interest in hybrid approaches that combine replay-based and architecture methods. These hybrid strategies appear promising for improving memory efficiency, scalability, and learning flexibility, which are crucial for advancing OCL.

By summarizing recent findings, this review gives a broad view of the current state of OCL, including main trends, important challenges, and possible future directions. We hope this work will become a valuable reference for researchers developing stronger, more efficient, and practical OCL methods.

\section*{Acknowledgments}
During the preparation of this work, the authors used ChatGPT~\cite{OpenAI2023} to improve the language. After using this tool, the authors reviewed and edited the content as needed and took full responsibility for the content of the publication.

\newpage

\appendix

\section{Newman's Guideline}\label{g}

\begin{enumerate}

\item \textbf{Develop Research Questions}: This step is the base of the review. It is about creating focused questions to guide the study. These questions should be detailed enough to find relevant studies but also wide enough to cover all important areas. Clear research questions decide the review’s limits and help make sure it gives useful and reliable results.

\item \textbf{Design Conceptual Framework}: Making a conceptual framework gives a structure for organizing and analyzing the literature. It shows the main ideas, variables, and their connections. This framework helps in choosing studies and ensures the analysis stays on track with the study’s goals. A good framework improves the review’s clarity and makes the findings easier to understand.

\item \textbf{Construct Selection Criteria}: The selection criteria set rules for including or excluding studies. These rules bring consistency to the process and remove bias, making sure only relevant and high-quality studies are part of the review. This step makes the results more trustworthy.

\item \textbf{Develop Search Strategy}: The search strategy is a plan for finding relevant studies from different databases. By using chosen keywords, Boolean operators, and clear criteria, this step ensures a complete collection of useful papers. A strong search strategy avoids missing important studies.

\item \textbf{Select Studies Using Selection Criteria}: In this step, we pick the studies that match the selection criteria. It helps to ensure that only the studies directly connected to the research questions are included. This step reduces confusion and keeps the review focused.

\item \textbf{Code Studies}: Coding means organizing the important information from each study into specific categories. This step makes the data easy to analyze and directly linked to the research questions. It helps in finding insights in a structured way.

\item \textbf{Assess the Quality of Studies}: After deciding which studies to include, we check their quality. This ensures the selected studies are good enough to provide reliable results. Poor-quality studies are excluded to keep the review rigorous.

\item \textbf{Synthesize Results of Individual Studies to Answer the Research Questions}: In this step, we combine the results of the selected studies to address the research questions. This involves finding patterns, similarities, or differences across studies and making conclusions based on these findings.

\item \textbf{Report Findings}: The last step is sharing the results in a clear and organized way. This includes summaries of the conclusions, study limitations, and suggestions for future research. This step provides valuable knowledge for the field.

\end{enumerate}

\newpage

\section{Components Tables}\label{componentsappendix}

\begin{table}[ht!]
   
    \centering  
    \caption{Mapping between Approaches and their Components (1/2).}  
    \label{componentsappendix1}  
    \begin{tabular}{c}  
        \includegraphics[width=1\linewidth]{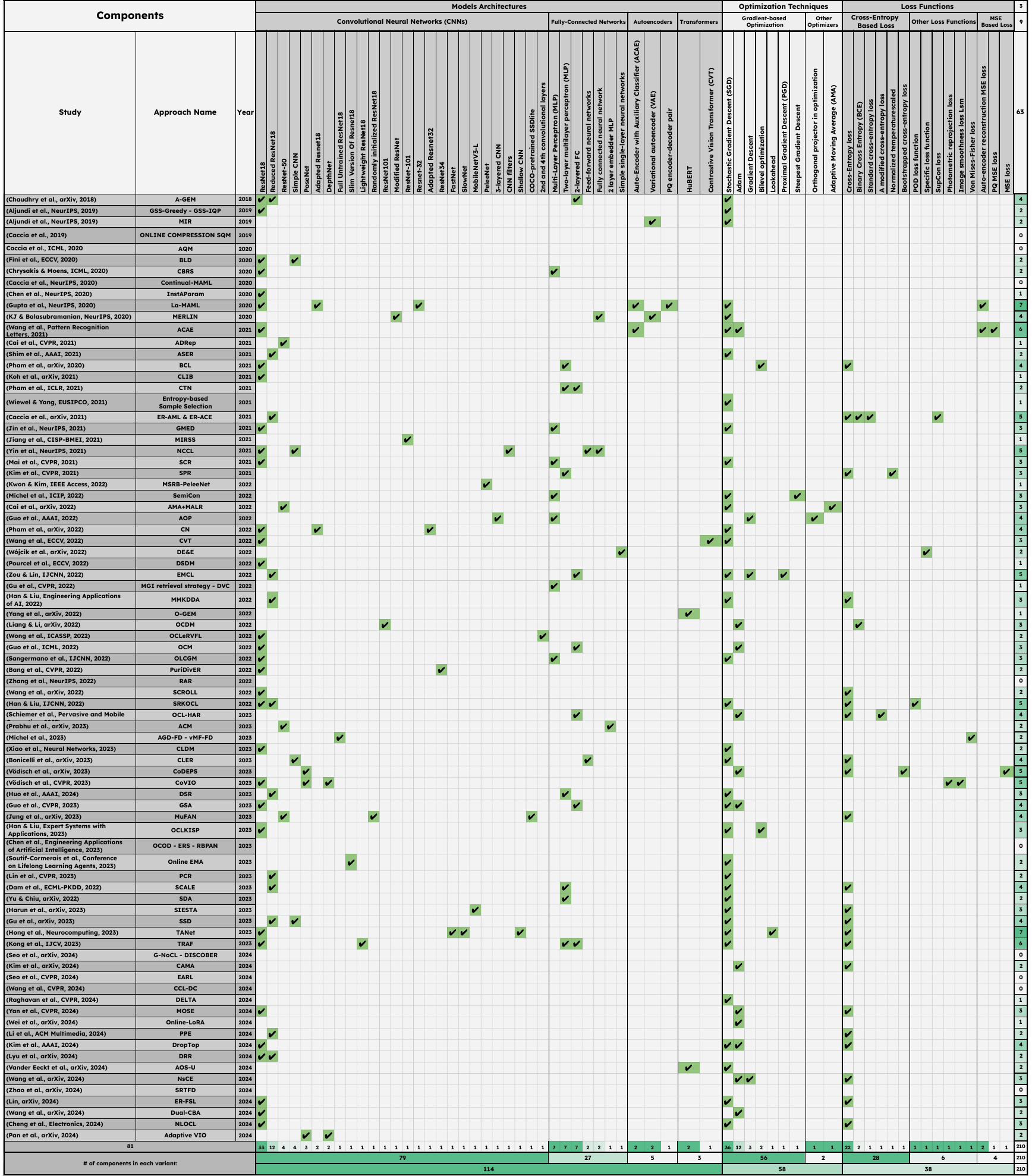} \\
    \end{tabular}  
     \vfill 
\end{table}

\newpage

\begin{table}[ht!]
    \centering
    \caption{Mapping between Approaches and their Components (2/2).}
    \label{componentsappendix2}
    \begin{tabular}{c}
        \includegraphics[width=1\linewidth]{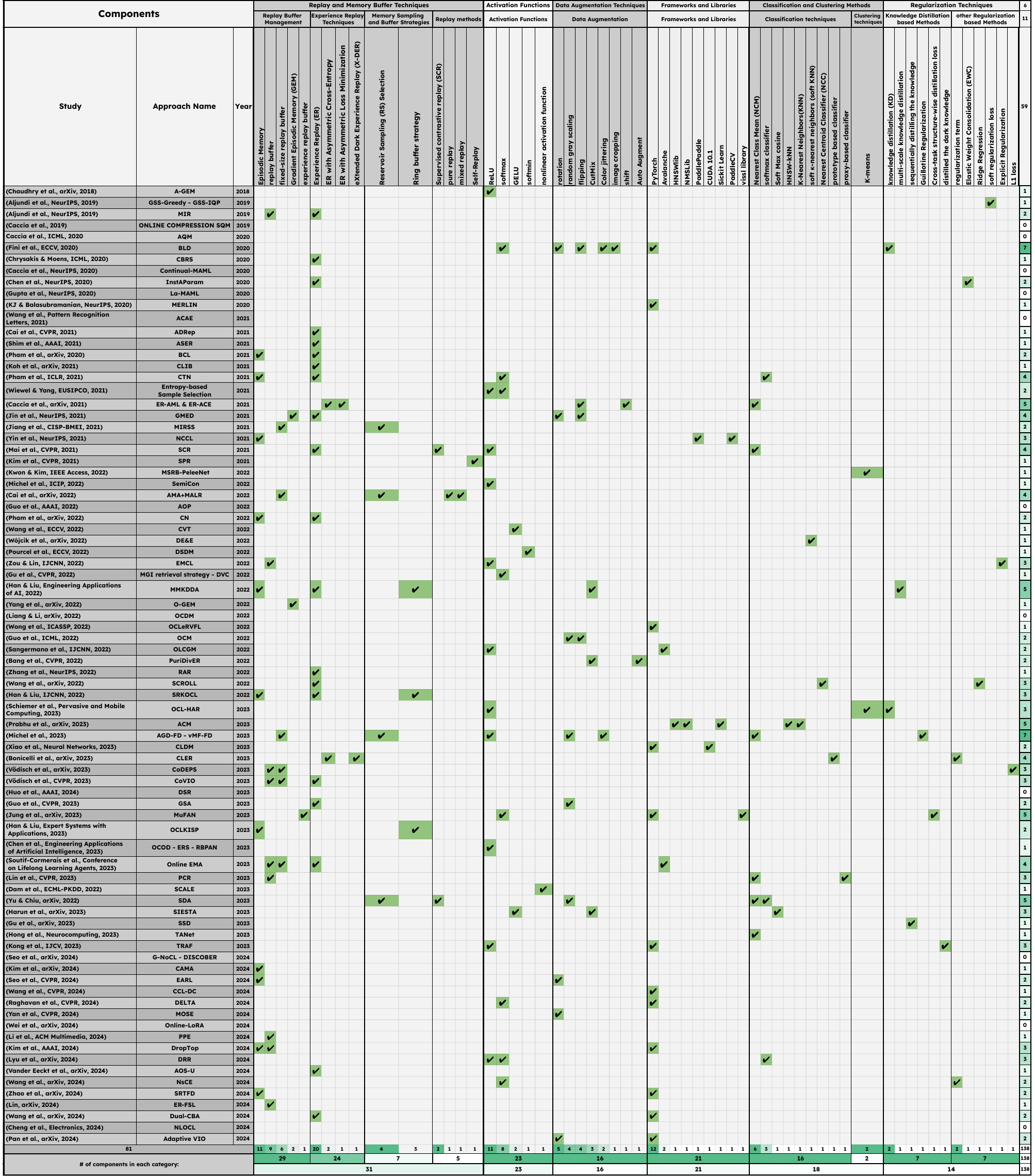} \\
    \end{tabular}
    \vfill
\end{table}

\newpage

\section{Features Table}\label{featureappendix1}

\begin{table}[ht!]
    \centering
    \caption{Mapping between Approaches and their Features.}
    \label{featureappendix}
    \begin{tabular}{c}
        \includegraphics[width=0.85\linewidth]{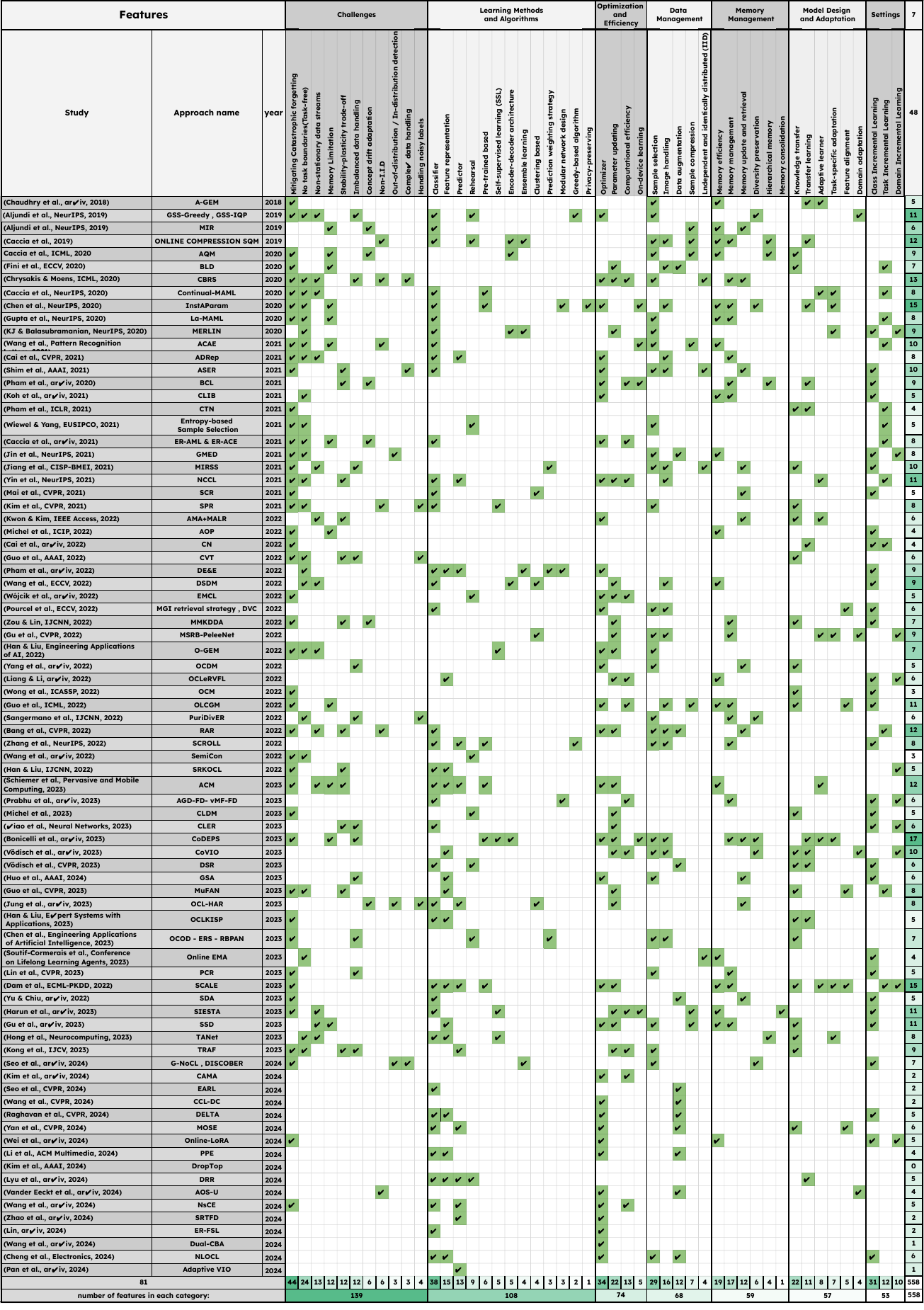} \\
    \end{tabular}
    \vfill
\end{table}

\newpage

\section{Quality Attributes Table}\label{qualityappendix1}

\begin{table}[ht!]
    \centering
    \caption{Mapping between Approaches and their Quality attributes.}
    \label{qualityappendix}
    \begin{tabular}{c}
        \includegraphics[width=1\linewidth]{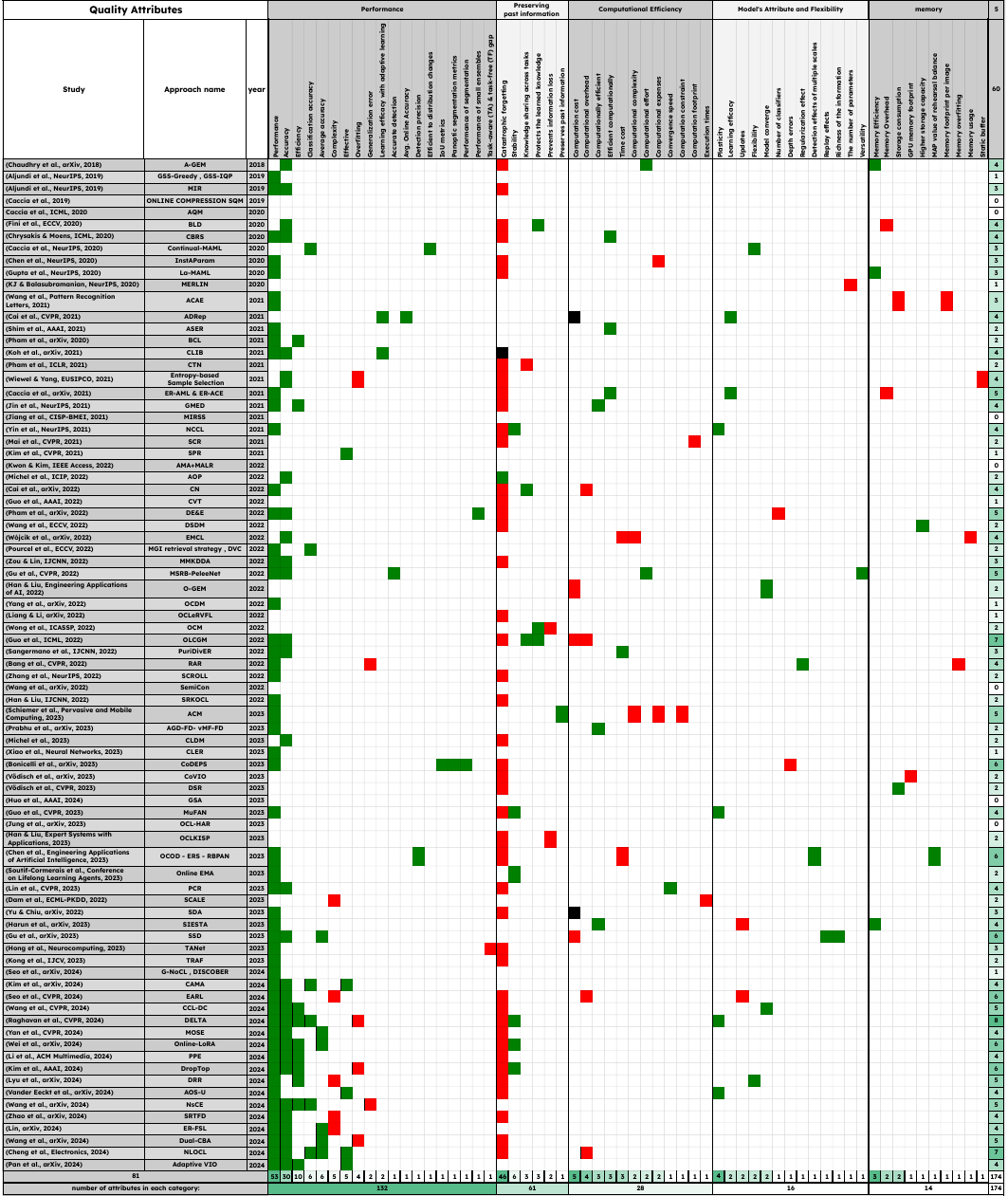} \\
    \end{tabular}
    \vfill
\end{table}

\newpage

\section{Dataset Table}\label{datasetappendix}

\begin{table}[ht!]
    \centering
    \caption{Mapping between Approaches and Datasets.}
    \label{}
    \begin{tabular}{c}
        \includegraphics[width=1\linewidth]{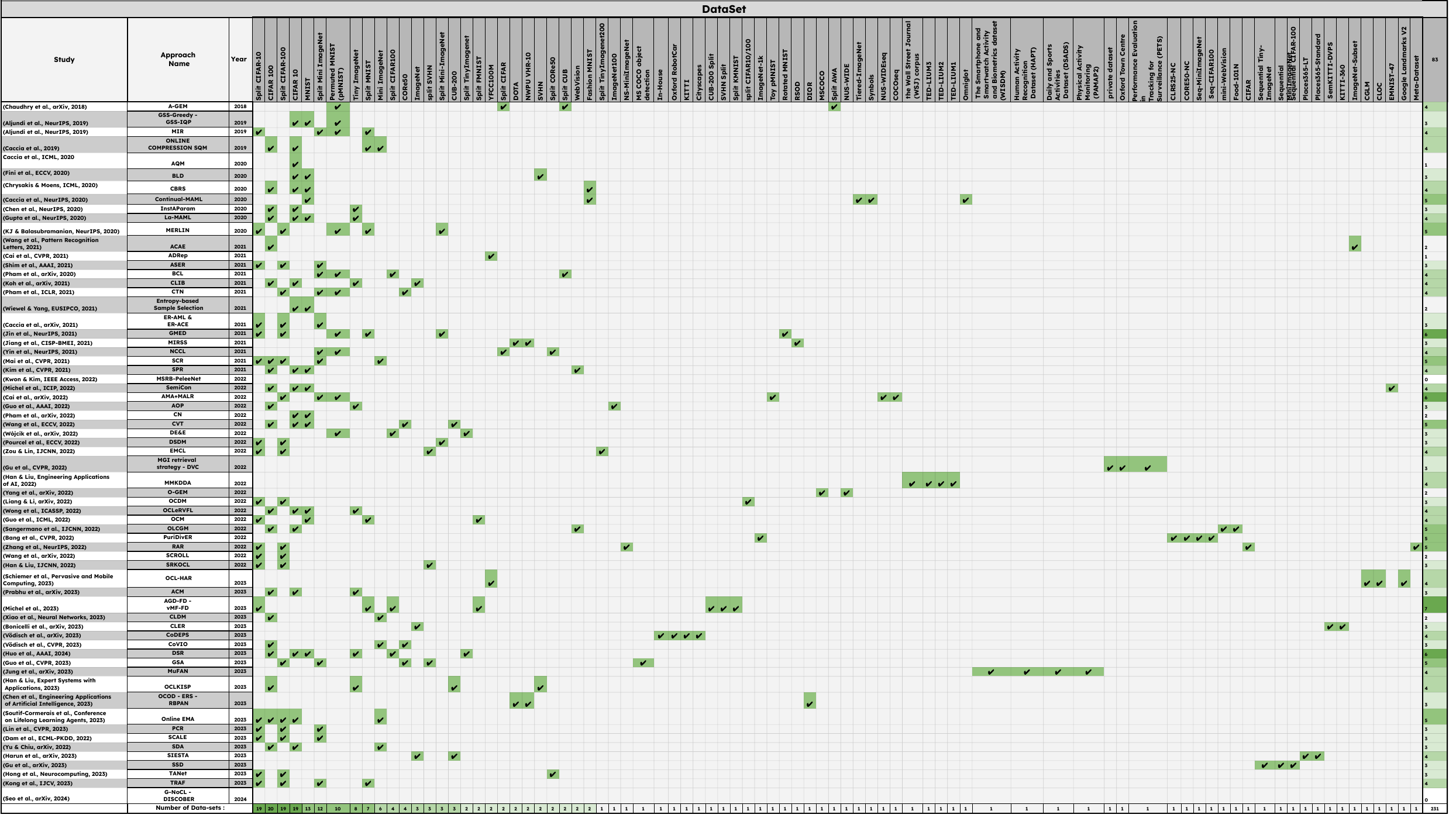}
    \end{tabular}
    \vfill
\end{table}

\end{document}